\documentclass[review]{elsarticle}

\journal{Journal of \LaTeX\ Templates}

\usepackage{algorithm}
\usepackage{algorithmic}
\usepackage{color}
\usepackage{multirow}
\usepackage{supertabular}
\usepackage{longtable}
\usepackage{subfigure}
\usepackage{caption}
\usepackage{amsthm}
\usepackage{amsmath}
\usepackage{amssymb}
\usepackage{hyperref}
\usepackage{url}
\usepackage[normalem]{ulem}

\begin{document}

\begin{frontmatter}
\title{General Method for Solving Four Types of SAT Problems}
%\title{A unified solving framework for different types of SAT problems based on reinforcement learning}
%\title{Elsevier \LaTeX\ template\tnoteref{mytitlenote}}
%\tnotetext[mytitlenote]{Fully documented templates are available in the elsarticle package on %\href{http://www.ctan.org/tex-archive/macros/latex/contrib/elsarticle}{CTAN}.}

%% Group authors per affiliation:
%\author{Elsevier\fnref{myfootnote}}
%\address{Radarweg 29, Amsterdam}
%\fntext[myfootnote]{Since 1880.}

%% or include affiliations in footnotes:
\author[mymainaddress]{Anqi Li}
\ead{lianqi20@mails.ucas.ac.cn}
%\ead[url]{www.elsevier.com}

\author[mymainaddress]{Congying Han\corref{mycorrespondingauthor}}
\cortext[mycorrespondingauthor]{Corresponding author}
\ead{hancy@ucas.ac.cn}

\author[mymainaddress]{Tiande Guo}
\ead{tdguo@ucas.ac.cn}

\author[mymainaddress]{Haoran Li}
\ead{lihaoran21@mails.ucas.ac.cn}

\author[mymainaddress]{Bonan Li}
\ead{libonan16@mails.ucas.ac.cn}

\address[mymainaddress]{School of Mathematical Sciences, University of Chinese Academy of Sciences, Shijingshan District, Beijing, China}

%\address[mymainaddress]{1600 John F Kennedy Boulevard, Philadelphia}
%\address[mysecondaryaddress]{360 Park Avenue South, New York}

\begin{abstract}
Existing methods provide varying algorithms for different types of Boolean satisfiability problems  (SAT), lacking a general solution framework. Accordingly, this study proposes a unified framework DCSAT based on integer programming and reinforcement learning (RL) algorithm to solve different types of SAT problems such as MaxSAT, Weighted MaxSAT, PMS, WPMS. Specifically, we first construct a consolidated integer programming representation for four types of SAT problems by adjusting objective function coefficients. Secondly, we construct an appropriate reinforcement learning models based on the 0-1 integer programming for SAT problems. Based on the binary tree search structure, we apply the Monte Carlo tree search (MCTS) method on SAT problems. Finally, we prove that this method can find all optimal Boolean assignments based on Wiener-khinchin law of large Numbers. We experimentally verify that this paradigm can prune the unnecessary search space to find the optimal Boolean assignments for the problem. Furthermore, the proposed method can provide diverse labels for supervised learning methods for SAT problems.
\end{abstract}

\begin{keyword}
%\texttt
{Boolean satisfiability problems}\sep {reinforcement learning}
\MSC[2010] 00-01\sep  99-00
\end{keyword}

\end{frontmatter}

%\linenumbers

\section{Introduction}
Boolean satisfiability problem (SAT) is a very classical combinatorial optimization problem. There are many variants depending on the number of variables and the weight of clauses, such as MaxSAT, Weighted MaxSAT, Partial MaxSAT (PMS), Weighted Partial MaxSAT (WPMS). In recent years, many scholars have devoted themselves to this problem~\cite{MCTS_SAT_relatedwork_1,MCTS_SAT_relatedwork_2,MCTS_SAT_relatedwork_3}. However, most of the methods are applicable to a specific type of problems. There is still lack of a general and efficient unified solution framework.

\begin{figure*}[t]
\centering
\includegraphics[width=1.00\columnwidth]{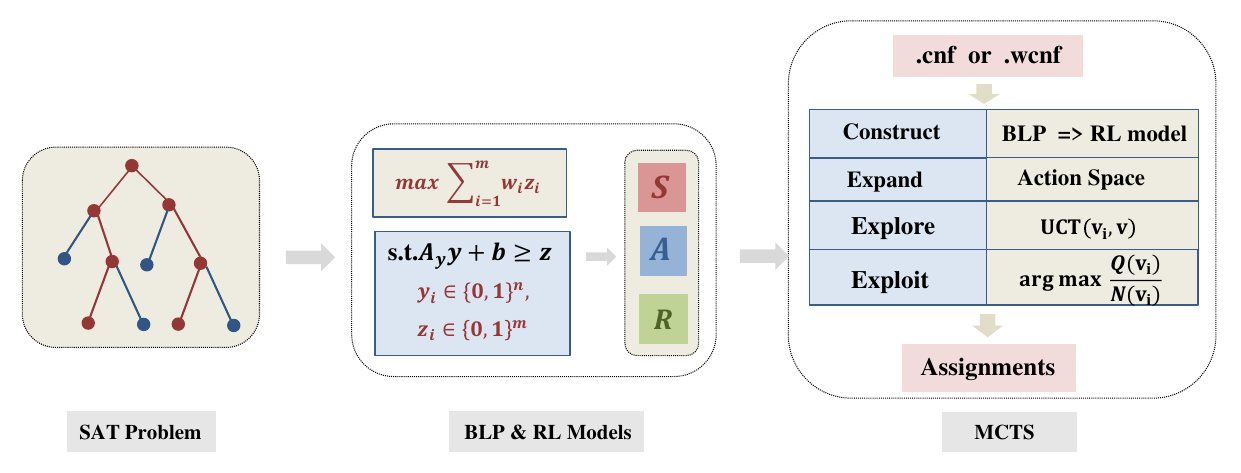}
\caption{Overview of the methodological framework. Firstly, we transform the binary tree of the general SAT problem into a BLP model. Based on this, we design appropriate reinforcement learning models. Then we use the MCTS method to find all the optimal Boolean assignments. Finally, we carry out in-depth theoretical analysis to prove the optimality and completeness.}
\label{fig:overview}
\end{figure*}

Motivated by these observations, we propose to analyze and improve the Boolean satisfiability problem with the Monte Carlo tree search (MCTS), further pushing forward the frontier of the Boolean satisfiability problem in a general way. This study focuses on four core aspects: (1) transformed into a general binary integer programming framework, (2) propose an appropriate reinforcement learning (RL) model, (3) giving a unified MCTS algorithm, and (4) providing thorough theory for the optimality and complexity, as shown in Figure~\ref{fig:overview}.

Constructing a unified representation for different SAT problems is the prerequisite for proposing a unified solution framework. Given that Boolean variables take true or false values, we spontaneously consider binary linear programming (BLP). First, we reagard Boolean variables as binary variables. On this basis, we introduce binary variables for each clause to measure whether the clause is satisfied. The optimization objective is to maximize the weighted sum of satisfied clauses. When the clause is not satisfied, the corresponding binary variable can only take $0$. On the contrary, the binary variable corresponding to the clause can take $1$ under the goal of maximization. We generalize the weighted and partial SAT problems into a unified framework through different weights.

Subsequently, we transform the binary linear programming framework of SAT problem into a reinforcement-learning model. Following the simplex tableaux structure of linear programming, we propose SAT tableaux as the state representation of our reinforcement learning model. In addition, we set the action space to all feasible Boolean variable assignments in the current state. And after each action selection, we immediately update the action space to the set of unselected actions. Because the optimal solution of the BLP problem is equivalent to the optimal Boolean assignment of the four classes of SAT problems. We set the reward as the objective function of the BLP. 

Furthermore, we naturally consider application of reinforcement learning method, based on the unified BLP framework and RL models for different SAT problems. Considering the binary tree structure of SAT problems, we undertake the Monte Carlo tree search method. We conduct exploratory estimation on all feasible action spaces and select the most appropriate action assignment according to the evaluation results. In this way, we ensure the optimality of Boolean variable assignment.

Consequently, we prove the optimality and completeness of the MCTS-based assignment method. The MCTS framework can acquire all optimal assignments according to Wiener-khinchin law of large Numbers. In particular, the MCTS method can find the optimal assignment when explorations become infinite. Additionally, it can find all the different pivot paths when executions approach infinity.

Given four intentions hereinabove, we develop a highly efficient reinforcement learning framework, which provides all optimal assignments for gemeral SAT problems. Moreover, we can label massive instances with little overhead. Substantial experiments on the Max-SAT-2016 and SATLIB benchmark demonstrate the effectiveness. Notably, our method can solve extensive instances compared with sat4j and other advanced methods. 

Our main contributions are summarized as follows: 
\begin{itemize}
\item Construct a genaral framework for solving varying SAT problems.
\item Propose a method to determine all the optimal Boolean assignments.
\item Give comprehensive theory for the optimality and complexity of the MCTS-based assignment method.
\end{itemize} 

\section{Related Work}
\paragraph{\textbf{SAT Solving Based on Machine Learning}}
In recent years, machine learning methods are widely used in combinatorial optimization problems~\cite{Optimize_7,Optimize_8,MCTSrule}, such as backpack~\cite{Optimize_4}, TSP~\cite{yvchen,Optimize_5,UseMCTS_TSP_1}, and P-median problem~\cite{Optimize_6}. As the first NP-complete problem to be proven, the SAT problem has also aroused the interest of many researchers~\cite{IandC_1,IandC_2,IandC_3,IandC_4,IandC_5,IandC_6,IandC_7,IandC_8,IandC_9}. Some studies have proposed to use neural networks to approximate the optimal heuristic algorithm in SAT solvers~\cite{MCTS_SAT_relatedwork_1,MCTS_SAT_relatedwork_2}. However, the performance is easily limited by the heuristic algorithm, and it is difficult to ensure the optimality~\cite{MCTS_SAT_relatedwork_3}. Selsam et al. ~\cite{MCTS_SAT_relatedwork_4} proposed the NeuroSAT, which transformed the SAT solving problem into a binary classification problem. Based on this, Selsam et al. ~\cite{MCTS_SAT_relatedwork_5} proposed NeuroCore, which uses the same architecture to guide variable branching between high-performance solvers (e.g., MiniSAT, Glucose). Amizadeh et al. ~\cite{MCTS_SAT_relatedwork_3} considered the problem of solving the Circuit-SAT problem. Kurin et al. ~\cite{MCTS_SAT_relatedwork_7} applied DQN to the SAT problem and trained a GNN model to predict the branch heuristic in the MiniSAT solver. Similarly, a reinforcement learning framework is used to train GNN models to learn a local-search heuristic algorithm~\cite{MCTS_SAT_relatedwork_8}. Vaezipor et al. ~\cite{MCTS_SAT_relatedwork_9} improved the \#SAT solver based on the Evolution Strategy. However, these methods are often only applicable to a specific type of SAT problem, and lack a general framework~\cite{MCTS_SAT_relatedwork_10}.

\paragraph{\textbf{Combinatorial Optimization Methods Based on MCTS}} Since the appearance of AlphaGo series~\cite{AlphaGo,AlphaGoZero}, reinforcement learning represented by MCTS has been widely used in many scenarios~\cite{IandC_RL_1,IandC_MCTS_1,UseRL_1,UseRL_2,UseRL_3,UseRL_4}. A perspective is to design a unified framework is applicable to different kinds of combination optimization problem~\cite{UseMCTS_manyCo_1,UseMCTS_manyCo_2}. In addition, we can also design suitable algorithms for some specific combinatorial optimization problems, such as the traveling salesman problem~\cite{UseMCTS_TSP_1,UseMCTS_TSP_2,UseMCTS_TSP_3,UseMCTS_TSP_4} and Boolean satisfiability problem~\cite{UseMCTS_SAT_1,UseMCTS_SAT_2,UseMCTS_SAT_3,UseMCTS_SAT_4}. In this paper, we adopt the latter idea to study the SAT problem.

\section{New Mathematical Formulation for Four types of SAT Problems}
\subsection{Definition and Classification}
%Boolean Satisfiability Problem is whether a set of truth assignments can be found to make the assignment of Boolean expressions true.
In Boolean satisfiability problems, Boolean expressions consist of conjunctions of clauses, and clauses are composed of disjunctive literals. Given a set of Boolean variables $\{x_1,x_2,...,x_n\}$, a literal is either the Boolean variable itself $\{x_i\}$ or its negation $\{\lnot x_i\}$, where $n$ represents the number of Boolean variables. A clause consists of disjunction of variables, i.e., $c_j=l_{j1} \lor l_{j2}\lor ...\lor l_{jn_j}$, where $n_j$ denotes the number of literals in clause $c_j$. A Conjunctive Normal Form(CNF) $\mathcal{F}$ is the conjunctive of several clauses, i.e. $\mathcal{F}=c_1 \land c_2 \land ... \land c_m$, where $m$ represents the number of clauses. A truth assignment is to assign a True or False value to each variable in a Boolean expression. The clause is satisfied when at least one literal in the clause is true. A CNF formula is satisfied if and only if all of clauses are true.

Given a CNF formula, SAT is a decision-making problem, to determine whether there is an assignment to satisfy $\mathcal{F}$. MaxSAT~\cite{IandC_1} is an extension of SAT that aims to find an assignment that satisfies the maximum number of clauses. Given a weight for each clause, the optimization goal of Weighted MaxSAT is to maximize the total weighted of satisfied clauses. Another way to extend this is to divide the clauses into hard and soft. The hard clause is mandatory, while the soft clause does not make a hard requirement. PMS tends to satisfy as many soft clauses as possible. WPMS is a generalization of PMS that assigns a non-negative weight to each soft clause. WPMS aims to maximize the total weight of satisfied soft clauses. In the following, we propose a general framework that applies to the four types of problems described above.

\begin{figure*}[hbtp]
\centering
\includegraphics[width=1.00\columnwidth]{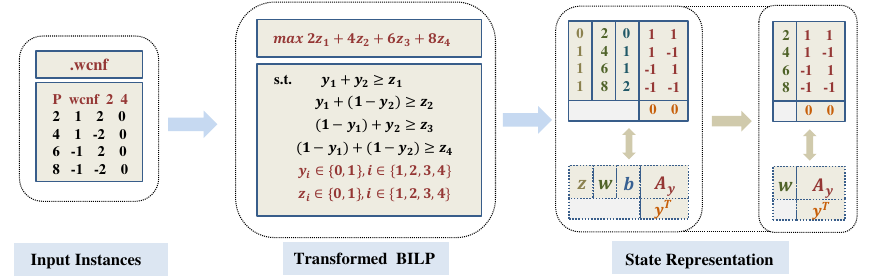}
\caption{Proposed SAT tableaux for SAT problems mentioned above. The subgraph on the left shows the input instance. Middle subgraph transforms a SAT instance into a BLP. In the subgraph on the right, the state representation of the SAT problem is proposed based on the simplex tableaux, and the SAT tableaux is obtained by reducing unnecessary coefficients. }
\label{fig:BIP}
\end{figure*}

\subsection{Description in BLP}
Four SAT problems mentioned above can be reduced to the form of binary integer linear programming. Each Boolean variable in $\{x_1,x_2,...,x_n\}$ corresponds to a binary variable $\{y_1,y_2,...,y_n\}$ respectively. In the same way, each clause in $\{c_1,c_2,...,c_m\}$ corresponds to a binary variable $\{z_1,z_2,...,z_m\}$. For each literal, the negation of the variable $x_i$ corresponds to $1-x_i$. The clause consists of disjunction of literals, which corresponds to the sum of transformed binary variables. A clause is satisfied when at least one of its corresponding literals is true, and vice versae. Because the binary variable corresponding to a clause can be true only if at least one literal is true (the clause satisfies), as Figure~\ref{fig:BIP} shows. The objective function is the weighted sum of satisfying clauses. Considering that the wcnf file format requires weights of hard clause always greater than the summation of weighted violated soft clauses in solution, we set the following objective coefficients for above four kinds of problems. (1) The weights of MaxSAT clauses are set to 1. (2) Weights of clauses are given values in the input instance for Weighted MaxSAT problem; (3) Weights of soft clauses are set to 1 and weights of hard clauses are given values in the input instance for PMS problems; (4) Weight of soft clauses and hard clauses in WPMS are given values in the input instance. This way set a unified BLP framework for the four kinds of SAT problems while ensuring that the optimal solution of the problem remains unchanged. 

In addition, following the simplex tableaux of linear programming, we set up a similar structure for the BLP of the SAT problem, which we call the SAT tableaux. SAT tableaux can save the complete information in the solution process, and it is also the best representation of the solution state. The binary variable $z$ corresponds to whether each clause is satisfied. The binary variable $y$ corresponds to the assignment of each variable, with $1$ representing True and $0$ representing False. $w$ represents the objective coefficients of BLP. Each component value of $b$ is the summation of constants in constraints. $A_y$ is the coefficient matrix of binary variables $\{y_1,y_2,...,y_n\}$ in BLP. Consider that $b$ can be derived from the negation of Boolean variables. At the same time, the satisfiable clauses $z$ can also be deduced when variable assignments are substituted in. Therefore, the redundant information $b$ and $z$ are subtracted when constructing SAT tableaux.

\subsection{RL Representation} 
We need to construct a unified reinforcement learning model for different types of SAT problems, such as MaxSAT, Weighted MaxSAT, PMS, WPMS. On the one hand, it ensures the generality of our method. On the other hand, it is the premise of using reinforcement learning methods. We follow the SAT tableaux proposed in previous section for state representation. 

The action space is the set of actions that can be performed in the next step. To ensure that the algorithm can find the optimal solution, we set the action space to the set of all possible actions to perform.
\begin{equation}
\label{eq:action_init}
A_{init}=\{y_1=1,y_1=0,y_2=1,y_2=0,...,y_n=1,y_n=0\}.
\end{equation}
Each time an action $y_i=1$ or $y_i=0$  is performed, we update the current action space $A$. The updated action space is obtained by deleting actions corresponding to the variable that have already been performed from the current action space. The action space becomes:
\begin{equation}
\label{eq:action_1}
A=A-\{y_i=0, y_i=1\}.
\end{equation}

Based on the BLP model we constructed for the four classes of SAT problems, finding the optimal solution to the SAT problem is equivalent to finding the optimal value of this BLP problem. Therefore, for the construction of the reward function, we adopt the most direct way. The reward function is set to the optimal value of the objective function. In this way, finding the optimal solution to these four classes of SAT problems is equivalent to maximizing the reward of the current reinforcement learning model. Our reward function takes the following form:
\begin{equation}
\label{eq:reward_3}
R=c^T y^*.
\end{equation}
For other feasible reward function settings, we conduct experimental verification in Section 6.4 and find that the current reward is the best.

\section{Proposed RL Algorithm}
\begin{figure*}[t]
\centering
\includegraphics[width=1.0\columnwidth]{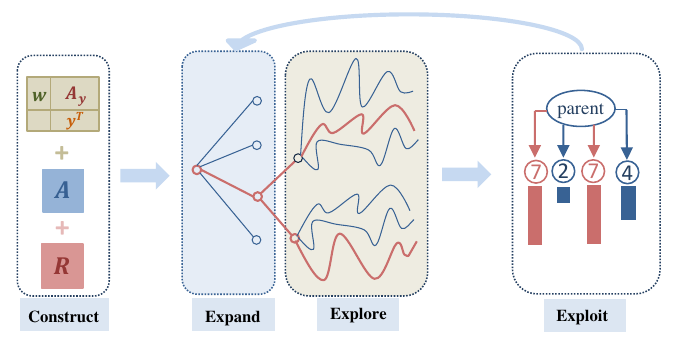} 
\caption{Algorithm flow diagram of the MCTS-based Boolean assignment method. }
\label{fig:flow_diagram}
\end{figure*}

MaxSAT problem is a sequential decision problem. Considering that the search space has a natural binary tree structure, we employ tree-based search based reinforcement learning methods. Monte Carlo tree search algorithm evaluates the future reward of each assignment through its prior exploration. Ultimately, we choose the action with the greatest expected reward, which guarantees the highest objective function value. The entire process is shown as Figure~\ref{fig:flow_diagram}, and the overall algorithm is shown in Algorithm~\ref{alg:MCTS_assignment}.

\begin{algorithm}[H]
\renewcommand{\algorithmicrequire}{\textbf{Input:}}
\renewcommand{\algorithmicensure}{\textbf{Output:}}
\caption{the MCTS assignment($I$,$N_{explore}$)}
\label{alg:MCTS_assignment}
\algsetup{linenosize=\tiny} \scriptsize
\begin{algorithmic}[1]
    \REQUIRE maximum number of explorations $N_{explore}$ and the input file $I$
    \ENSURE the optimal assignment $a^*$

    \STATE $s,A_s$=Transform($I$)
    \STATE create initial root node $v=null$
    \WHILE{$v$ is not terminal}
        \FOR{$a \in A_s$}
            \STATE $s^\prime=$ Transition($s$,$a$)
            \STATE $v^{'}=$ CreatNode($s^\prime,a,v$)
            \STATE $r=$ RewardEpisode($s,a,A_s$)
            \STATE BackUp$(v^\prime,r)$
        \ENDFOR 
 
        \FOR{$i \in {\lbrace{ |A_s|+1,|A_s|+2,\dots,N_{explore}}\rbrace} $}
            \STATE $tag=0$
            \STATE $v_{next},a=$ BestChild($v,tag$)
            \STATE $r=$ RewardEpisode($s,a,A_s$)
            \STATE BackUp$(v_{next},r)$
        \ENDFOR
        \STATE $tag=1$
        \STATE $v_{next},a=$ BestChild($v,tag$)
        \STATE $v = v_{next}$
        \STATE $s = v.state$
        \STATE $A_s$= search all possible assignments of $s$ to get the action space 
    \ENDWHILE
    \STATE get optimal 
    \end{algorithmic}
\end{algorithm}

\subsection{MCTS-based Boolean Assignment Method}
\paragraph{\textbf{Construct}} 
The premise of applying the MCTS method is to transform the problem into a reinforcement learning model. We follow the representation given in the previous section for extraction and model transformation. SAT tableaux are given as the state representation based on the BLP representation of the problem. We use all possible assignments to the current state as our action space. In addition, the reward is set to the obejective value of the current solution. Algorithm~\ref{alg:Transform} presents the pseudocode to transform cnf or wcnf files into state and action spaces.
% We use all possible assignments to the current state \textcolor{red}{(or the state after unit propagation)} as our action space. According to the different emphasis on the solution process and results, we adopt the four models proposed above.

\begin{algorithm}[H]
\renewcommand{\algorithmicrequire}{\textbf{Input:}}
\renewcommand{\algorithmicensure}{\textbf{Output:}}
\caption{Transform($I$)}
\label{alg:Transform}
\algsetup{linenosize=\tiny} \scriptsize
\begin{algorithmic}[1]
    \REQUIRE the input file $I$
    \ENSURE the transformed state $s$ and action space $A_s$
    \STATE $w,A_y,b=$ read the input file $I$
    %\textcolor{red}{\STATE $s=$ perform unit propagation and update $s$}
    \STATE set initial value $y=+\infty$
    \STATE $s$= concat $w,A_y,b,y$
    \STATE $A_s$= search all possible assignments of $s$ to get the action space 
    \end{algorithmic}
\end{algorithm}

\paragraph{\textbf{Expand}} 
In the expansion stage, we first need to identify the set $U$ of all unassigned variables in current state. To expand all nodes, we need to select all possible assignments $A_{cur}$ for the unassigned variable once. 
\begin{equation}
\label{eq:equation_expansion_1}
A_{cur}={\lbrace y_i=0, y_i=1 \vert i \in U \rbrace}.
\end{equation}
We use $I$ to represent the set of all selected variables. We start with $I=\emptyset$. After choosing a action $a_i$ for the unassigned variable, we renewal $I=I \cup {\lbrace i \rbrace}$ and update the state to retrieve unassigned variables. Moreover, we use a random strategy to generate episodes.
\begin{equation}
\label{eq:equation_expansion_2}
randomly\ select\ from\ {\lbrace a_i\in A_{cur} \vert i \notin I \rbrace}.
\end{equation}

\begin{algorithm}[H]
\renewcommand{\algorithmicrequire}{\textbf{Input:}}
\renewcommand{\algorithmicensure}{\textbf{Output:}}
\caption{Transition($s$,$a$)}
\label{alg:Transition}
\algsetup{linenosize=\tiny} \scriptsize
\begin{algorithmic}[1]
    \REQUIRE the current state $s$ and the action $a$
    \ENSURE the transformed state $s^\prime$
    
    \STATE $y^\prime=$ new assignment execute by $a$
    \STATE $s^\prime=$ update state $s$
    
    \end{algorithmic}
\end{algorithm}

\begin{algorithm}[H]
\renewcommand{\algorithmicrequire}{\textbf{Input:}}
\renewcommand{\algorithmicensure}{\textbf{Output:}}
\caption{CreatNode($s^\prime,a,v$)}
\label{alg:CreatNode}
\algsetup{linenosize=\tiny} \scriptsize
\begin{algorithmic}[1]
    \REQUIRE the state $s^\prime$, action $a$ and parent node $v$
    \ENSURE the created node $v^\prime$
    
    \STATE $v^\prime.state = s^\prime$
    \STATE $v^\prime.from\_action = a$
    \STATE $v^\prime.parent = v$
    \STATE $v^\prime.Q=0$
    \STATE $v^\prime.N=0$
    \end{algorithmic}
\end{algorithm}

We present in Algorithm~\ref{alg:Transition} the process of updating the state $s$ after taking an action $a$. We maintain the search process with a tree structure, as shown in Algorithm~\ref{alg:CreatNode}. Each branch according to perform an action. After the action is performed, we generate a child node. The child node holds the current state and the action is taken by the parent node. In other words, the current state is the new state obtained after the action is taken. %$Q(s,a)$ and $N(s,a)$ represent the current state action value function and the number of visits, respectively, which we will discuss in detail later. 
$Q$ and $N$ represent the current state-action value function and the number of visits, respectively, which we will discuss in detail later. 

After performing an action, we need to randomly construct episodes to calculate the reward. We compute the reward $r_{T}$ of an episode using the definitions in the reinforcement learning model. 
\begin{equation}
\label{eq:equation_3}
G_t=r_{T}
\end{equation}

Starting from the current state-action pair, the way we calculate reward is shown in Algorithm~\ref{alg:RewardEpisode}.
\begin{algorithm}[H]
\renewcommand{\algorithmicrequire}{\textbf{Input:}}
\renewcommand{\algorithmicensure}{\textbf{Output:}}
\caption{RewardEpisode($s,a,A_s$)}
\label{alg:RewardEpisode}
\algsetup{linenosize=\tiny} \scriptsize
\begin{algorithmic}[1]
    \REQUIRE the state $s$, action $a$ and action space $A_s$
    \ENSURE the reward $r$ from state $s$
    
    \STATE $episode=\emptyset$
    \WHILE{$s$ is not terminal}
        \STATE $a=$ randomly select from $A_s$
        \STATE $s^\prime=$ Transition($s$,$a$)
        % \STATE $s_1=$ perform unit propagation and update $s_1$
        \STATE $A_{s^\prime}=$ search all possible assignments of $s^\prime$ to get the action space 
        \STATE $episode.append(s,a,s^\prime)$
    \ENDWHILE
    
    \STATE $r=$ compute reward for $episode$
    \end{algorithmic}
\end{algorithm}

It is worth noting that only leaf nodes correspond to feasible solutions, and that each variable assignment contributes equally to the optimal solution. In addition, the strategy evaluation approach employs the empirical mean of rewards as the expectation.
\begin{equation}
\label{eq:equation_4}
Q_{\pi}(s,a)=\mathbb{E}_{\pi} [G_t \vert s_t=s,a_t=a], \  t\in \lbrace1,2,...,T\rbrace,
\end{equation} 
where $\left\{1,2,...,T \right\}$ denotes subscripts of the current episode.
The expanded node and its parent node are then retroactively updated with the rewards from the entire episode. We add $1$ to the count function $N$ and increase the current state-action value function $Q$ by the cumulative reward when we encounter state $s$ in an episode and perform action $a$. The backtracking procedure based on tree structure is implemented in Algorithm~\ref{alg:BackUp}.
\begin{equation}
\label{eq:equation_5}
N(s,a) \leftarrow N(s,a)+1.
\end{equation}
\begin{equation}
\label{eq:equation_6}
Q(s,a) \leftarrow Q(s,a)+G_t.
\end{equation}

\begin{algorithm}[H]
\renewcommand{\algorithmicrequire}{\textbf{Input:}}
\renewcommand{\algorithmicensure}{\textbf{Output:}}
\caption{BackUp($v,r$)}
\label{alg:BackUp}
\algsetup{linenosize=\tiny} \scriptsize
\begin{algorithmic}[1]
    \REQUIRE the current node $v$ and reward $r$
    
    \WHILE{$v$ is not $null$}
        \STATE $v.Q=v.Q+r$
        \STATE $v.N=v.N+1$
        \STATE $v=v.parent$
    \ENDWHILE
    \end{algorithmic}
\end{algorithm}

\paragraph{\textbf{Explore}} 
The exploration stage follows the expansion stage. We need to estimate the future payoff of the current assignment by performing $N_{explore}$ exploration of the previously expanded node. Inspired by the MCTS rule~\cite{MCTSrule}, we use soft max to relax the condition of the upper confidence
bounds applied to the trees (UCT) algorithm~\cite{UCT}. First, we use the UCT algorithm to estimate the current state-action value function.
\begin{equation}
\label{eq:equation_exploration_UCT}
Q(v^\prime)=\frac{v^\prime.Q}{v^\prime.N}+C\sqrt{\frac{2\ln{v.N}}{v^\prime.N}},\ v^\prime\in children\ of\ v.
\end{equation}
Where $v$ represents the current node, $Q$ represents the value of state-action function at the current node, and $N$ represents the number of visits to the current node. Then we relax the maximization by controlling $\alpha$, where $\alpha \in \mathbb{R}$ and $\alpha \in [0,1]$. Actions are randomly selected for execution from the set larger than soft max.
\begin{equation}
\label{eq:equation_exploration_Esoftmax}
E_{soft}=(1-\alpha)\mathop{\min}\limits_{v^\prime \in children\ of\ v}{Q(v^\prime)}+\alpha\mathop{\max}\limits_{v^\prime \in children\ of\ v}{Q(v^\prime)}
%E_{soft}=\mathop{\min}\limits_{v^\prime \in children\ of\ v}{Q(v^\prime)}+\alpha\left(\mathop{\max}\limits_{v^\prime \in children\ of\ v}{Q(v^\prime)}-\mathop{\min}\limits_{v^\prime \in children\ of\ v}{Q(v^\prime)} \right)
\end{equation}

\begin{equation}
\label{eq:equation_exploration_select}
v_{next}=random \ select \ from \ \lbrace {v^\prime} \vert {Q(v^\prime)}\ge E_{soft},v^\prime \in children\ of\ v \rbrace
\end{equation}
This part is the execution process of Algorithm~\ref{alg:BestChild} when $tag=0$. Subsequently, we continue to generate episodes in the random manner given by Formula (\ref{eq:equation_expansion_2}) and the rewards are calculated in the same way as in Algorithm~\ref{alg:RewardEpisode}. We calculate the rewards according to the previously defined reinforcement learning model. The existing tree structure information is updated according to Formulas (\ref{eq:equation_5}) and (\ref{eq:equation_6}), as Algorithm~\ref{alg:BackUp} shows. When explorations reaches $N_{explore}$, the exploration ends. Evaluation of each assignment under $N_{explore}$ exploration is given by the state-action value function.

\begin{algorithm}[H]
\renewcommand{\algorithmicrequire}{\textbf{Input:}}
\renewcommand{\algorithmicensure}{\textbf{Output:}}
\caption{BestChild($v,tag$)}
\label{alg:BestChild}
\algsetup{linenosize=\tiny} \scriptsize
\begin{algorithmic}[1]
    \REQUIRE the current node $v$ and bool mark $tag$ 
    \ENSURE the selected action $a$ and node $v_{next}$

    \IF{tag=0}
        \STATE $Q(v^\prime)=\frac{v^\prime.Q}{v^\prime.N}+C\sqrt{\frac{2\ln{v.N}}{v^\prime.N}},\ v^\prime\in children\ of\ v$
        \STATE $E_{soft}=\mathop{\min}\limits_{v^\prime \in children\ of\ v}{Q(v^\prime)}+\alpha\left(\mathop{\max}\limits_{v^\prime \in children\ of\ v}{Q(v^\prime)}-\mathop{\min}\limits_{v^\prime \in children\ of\ v}{Q(v^\prime)} \right)$
        \STATE $v_{next}=random \ select \ from \ \lbrace {v^\prime} \vert {Q(v^\prime)}\ge E_{soft},v^\prime \in children\ of\ v \rbrace$
    \ELSE
        \STATE $Q(v^\prime)=\frac{v^\prime.Q}{v^\prime.N}$
        \STATE $v_{next}=\mathop{\arg\max}\limits_{v^{'} \in children\ of\ v} Q(v^\prime)$
    \ENDIF
    \STATE $a = v_{next}.from\_action$
    \end{algorithmic}
\end{algorithm}

\paragraph{\textbf{Exploit}}
In the exploitation phase, we select the optimal assignment based on the evaluation results of the exploration phase. The $N_{explore} $ exploration has been completed. Furthermore, the state-action value function effectively reflects the future reward after exploration. Assignments with large state-action value functions correspond to large rewards. Therefore, we choose the assignment as follows:
\begin{equation}
\label{eq:equation_exploitation_1}
v^{*}=\mathop{\arg\max}\limits_{v^{''}\in children\ of\ v}\frac{Q(v^{''})}{N(v^{''})}.
\end{equation}
\begin{equation}
\label{eq:equation_exploitation_2}
a^{*}=\mathop{\arg\max}\limits_{a \in A}Q(s,a),
\end{equation}
The above process is the case of Algorithm~\ref{alg:BestChild} when $tag=1$. We update the status after selecting the optimal assignment. The above process is completed until all variables are assigned. Otherwise we return to the expansion stage and repeat the cycle.

\subsection{Extracting Multiple Boolean Assignments}
Optimal solutions to the MaxSAT problem is the Boolean assignment that maximizes objective values. The assignment of variables is not necessarily unique while maximizing the objective values. Considering the randomness of action selection in the exploration phase, the action of exploration is random when we maximize the objective value. Intuitively, this MCTS-based approach can find different optimal solutions if it exists. Subsequently, we analysis the convergence property of our MCTS-based approach.

\section{Theoretical Analysis}
In this section, we present a detailed theoretical analysis of our framework in terms of optimality and completeness. First, we prove that the DCSAT that can perform optimal Boolean variable assignment at each step to find the optimal solution. It is further proved that our method can find all the optimal assignments when algorithm executions are sufficient.

\subsection{Optimality of the DCSAT}
We show that the DCSAT will find the optimal solution as explorations approximates infinity. Considering that the expectation of reward is affected by other episodes, we define the significance operator $Sig$ based on convolutional pooling. We then provide the full proof details.

\newtheorem{Definition}{Definition}[section]
\begin{Definition}[Rank Function]
    \label{def: rank func}
    Given probability space ${\Omega}$ and a sequence of random variables $\mathrm{RS} := \{ R_1, R_2, \dots, R_n \} \subset {\Omega}$, $\mathrm{RS_O} := \{ R_{(1)}, R_{(2)}, \dots, R_{(n)} \} $ is the order statistics sequence of $\mathrm{RS}$, where $R_{(i)}$ represents the $i$-th smallest random variable. Define the function $\operatorname{Rank}_{\mathrm{RS}}: {\Omega} \longrightarrow [n]$, where $[n] := \{ 1,2,\dots,n \}$, and $\operatorname{Rank}_{\mathrm{RS}}(R_i)$ is the index of $R_i$ in $\mathrm{RS_O}$.
\end{Definition}

% \begin{Definition}[Rank Function]
%     \label{def: rank func}
%     Let $\mathcal{X}=\{ X_1, X_2, \dots, X_n \}, [n] := \{ 1,2,\dots,n \}$, define the rank function $Rank: \mathcal{X}\Longrightarrow [n]$
    
%     %Given a sequence of random variables $\mathrm{RS} := \{ X_1, X_2, \dots, X_n \} \subset \mathcal{X}$, it is sorted with an ascending order to get the order statistics sequence $\mathrm{RS_O} := \{ X_{(1)}, X_{(2)}, \dots, X_{(n)} \} $, where $X_{(i)}$ represents the $i$-th smallest random variable. Here we define the function $\operatorname{Rank}_{\mathrm{RS}}: \mathcal{X} \longrightarrow [n]$, where $[n] := \{ 1,2,\dots,n \}$, and $\operatorname{Rank}_{\mathrm{RS}}(X_i)$ is the index of $X_i$ in $\mathrm{RS_O}$.
% \end{Definition}

\begin{Definition}[Significance Operator]
\label{Def1}
Given K groups of random variables sequences $\lbrace R^k_1,R^k_2,...,R^k_{n_k}\rbrace$, $k \in \lbrace 1,2,...,K \rbrace$, we define $\bar{R}^k$ and $R_{(n_k)}^k$ as the mean statistic and the maximum statistic of $k$-th sequence $\lbrace R^k_i\rbrace_{i=1}^{n_k}$, respectively:
$$\bar{R}^k = \frac{1}{n_k} \sum_{i=1}^{n_k} R_i^k,\ R_{(n_k)}^k = \max \{ R^k_1,R^k_2,...,R^k_{n_k} \},\ k \in \lbrace 1,2,...,K \rbrace.$$
For the mean statistics sequence $\mathrm{ES} := \{ \bar{R}^1,\bar{R}^2,\dots, \bar{R}^K \}$ and the maximum statistics sequence $\mathrm{MS} := \{ R_{(n_1)}^1,R_{(n_2)}^2,\dots, R_{(n_K)}^K \}$, define the significance operator $\operatorname{Sig}: \Omega \longrightarrow \Omega$ as:
\begin{equation}
    \operatorname{Sig}(\bar{R}^k)=
    \begin{cases}
    \begin{array}{r}
        \bar{R}^k,\ \operatorname{Rank}_{\mathrm{ES}}(\bar{R}^k) = \operatorname{Rank}_{\mathrm{MS}}(R_{(n_k)}^k)\\
        R_{(n_k)}^k,\ \operatorname{Rank}_{\mathrm{ES}}(\bar{R}^k) \neq \operatorname{Rank}_{\mathrm{MS}}(R_{(n_k)}^k).
    \end{array}
    \end{cases}
\end{equation}
\end{Definition}
%If the index of $\bar{X_k}$ in $\mathrm{ES_O}$ is equal to the index of $X_{(n_k)}^k$ in $\mathrm{MS_O}$, the operator $\operatorname{Sig}$ maps $\bar{X_k}$ to itself and otherwise, the operator $\operatorname{Sig}$ maps $\bar{X_k}$ to $X_{(n_k)}^k$.

\newtheorem{thm}{\bf Theorem}[section]
\begin{thm}\label{thm1}
If we set $\alpha$ in Formula (\ref{eq:equation_exploration_Esoftmax}) as zero, the DCSAT with significance operator $\operatorname{Sig}$ will converge to the optimal Boolean assignment as $N_{explore}$ goes to the infinity.
\end{thm}
\begin{proof}
In our reinforcement learning model, the reward is defined as the objective value of the transformed BLP problem. Therefore, maximizing the reward is equivalent to finding the optimal Boolean assignment for the four types of SAT problems.

To avoid confusion, we assume that the depth of root node is zero. Given a SAT problem, our model constructs a tree of depth $n$, where $n$ is the number of decision variables. 
Without loss of generality, we consider the state node $s$ of which the depth is $k$. Note that $k$ variables are assignmented in state $s$ and denote the feasible action space in $s$ as $\mathcal{A}_s:=\{ a_1, a_2,\dots, a_{2(n-k)} \}$ where each action corresponds to a feasible variable assignment. The child nodes of $s$ are $\{ s_1, s_2, \dots, s_{2(n-k)}\}$ of which the depth is $k+1$ and the transition functions are $\mathbb{P}(s_i | s, a_j) = \mathbb{I}_{\{ i=j \}},\ \forall i,j\in \{1,2,\dots,{2(n-k)}\}$, where $\mathbb{I}$ is the indicator function. 

We denote $N_{explore}$ as the number of explorations from $s$. The random variable $N_i$ is the number of explorations from $s$ to $s_i$ and $\sum_{i=1}^{2(n-k)} N_i = N_{explore}$. 

If we set $\alpha$ in Formula (\ref{eq:equation_exploration_Esoftmax}) as zero, the set of selected actions is
\begin{equation}
\label{eq:equation_thm1_1}
 \left\{ i|Q(s,a_i) \ge E_{soft} \right\} = \left\{ i|Q(s,a_i) \ge \min_{a_j\in\mathcal{A}_s} Q(s,a_j)\right\} = \mathcal{A}_s,
\end{equation}
which indicates that all possible variable assignment can be selected, i.e. $\mathbb{E}_{\pi}\left[N_i\right] > 0,\ \forall i\in \{1,2,\dots,2(n-k)\}$. In fact, our algorithm takes the exploration action by uniform policy from the set $\left\{ i|Q(s,a_i) \ge E_{soft} \right\}$, and then we have
\begin{equation}
\label{eq:equation_thm1_2}
\begin{aligned}
\mathbb{E}_{\pi}\left[N_i\right] &=  \sum_{i=1}^{|\mathcal{A}_s|} \mathbb{P}(s_i | s, a_j) \pi(a_j|s) N_{explore} \\
&= \pi(a_i|s) N_{explore} \\
&= \frac{N_{explore}}{|\left\{ i|Q(s,a_i) \ge E_{soft} \right\}|} \\
&= \frac{1}{2(n-k)}N_{explore} > 0, \qquad  \forall i\in \{1,2,\dots,2(n-k)\}.
\end{aligned}
\end{equation}

Algorithm \ref{alg:MCTS_assignment} has two phases of exploration. In the first phase, every child node of $s$ is visited once. And in the second phase, child nodes are explored according to the proposed DCSAT. 
We can conclude that as long as $N_{explore} \ge \frac{1}{\ln \left( 1 + \frac{1}{2(n-k)-1}\right)} \ln \left( \frac{1}{\delta_1} \right) +2(n-k)$, every child node can be accessed to in the second phase with probability at least $1-\delta_1$, i.e., 
\begin{equation}\label{state visited}
    \begin{aligned}
        &\quad\ \mathbb{P}\left( \text{state $s_i$ visited} \right)\\
        &=\mathbb{P}{\left( N_i > 0 \right)}\\
        &= 1 - \mathbb{P}{\left( N_i = 0 \right)}\\
        &= 1 - \left(1- \frac{1}{2(n-k)} \right)^{N_{explore}-2(n-k)} \\
        &\ge 1-\delta_1.
    \end{aligned}
\end{equation}

We denote random variable $R_j^{a_i}$ as the reward of taking action $a_i$ for the $j^{th}$ time. Given $2(n-k)$ groups of independent identically distributed random variables sequences $\left\{ R_j^{a_i} \right\}_{j=1}^{N_i},\ i\in \{1,2,\dots,2(n-k)\}$, we apply the significance operator $\operatorname{Sig}$ in Definition \ref{Def1} and we have $\left\{ \operatorname{Sig}(\bar{R}^{a_i})\right\}_{i=1}^{2(n-k)}$. Note that the reward random variables $\left\{ R_j^{a_i} \right\}_{j=1}^{N_i}$ are independent and identically distributed, and we can apply Wiener-khinchin Law of Large Numbers: 
\begin{equation}
    \label{eq: LLN}
    \lim_{N_{explore} \to \infty} \bar{R}^{a_i} = \lim_{N_{i} \to \infty} \bar{R}^{a_i} =\lim_{N_{i} \to \infty} \frac{1}{N_i} \sum_{j=1}^{N_i} {R}^{a_i}_j = \mathbb{E}\left[ {R}^{a_i}_j \right],\quad a.s.
\end{equation}
Note that $Q(s,a_i)$ is the same as $\bar{R}^{a_i}$ with respect to the definition of $Q$.

% For any pivot path $P = \{s^P_0, s^P_1, \dots, s^P_{|P|-1}\}$, we denote $d_\mathcal{A}^P$ as the dimension of the maximum action space along this path, i.e. $d_\mathcal{A}^P:= \max_{s^P\in P} |\mathcal{A}_{s^P}|$. 
% By recursion, the MCTS rule can explore pivot path P with probability at least $1-\delta_2$ when $N_{explore} \ge \frac{1}{\ln \left( 1 + \frac{1}{d_\mathcal{A}^P-1}\right)} \ln \left( \frac{1}{1-e^{\frac{1}{|P|} \ln\left( 1- \delta_2\right)} } \right) \approx O\left( \frac{1}{\ln \left( 1 + \frac{1}{d_\mathcal{A}^P-1}\right)}  \ln \left( \frac{|P|}{\delta_2} \right) \right)$ , i.e. 

% \begin{equation} \label{eq: access to each path}
%     \begin{aligned}
%         &\quad\ \mathbb{P}\left( \text{explore pivot path $P$} \right)\\
%         &= \mathbb{P} \left( \bigcap_{s^P\in P} \left\{ N_{s^P} >0 \right\} \right)\\
%         &= \prod_{s^P\in P} \mathbb{P} \left( N_{s^P} >0 \right)\\
%         &= \prod_{s^P\in P} 1 - \left( \frac{|\mathcal{A}_{s^P}|-1}{|\mathcal{A}_{s^P}|} \right)^{N_{explore}} \\
%         &\ge \left( 1 - \left( \frac{d_\mathcal{A}^P-1}{d_\mathcal{A}^P} \right)^{N_{explore}} \right)^{|P|}\\
%         &\ge 1-\delta_2.
%     \end{aligned}
% \end{equation}

Note that our tree model for SAT problems has three special structures: 1) the depth of each leaf node is the same; 2) the cardinality of the acion space of nodes of the same depth is the same; 3) the cardinality of the action space of a node is an arithmetic progression with respect to depth.
For any feasible assignment $AST = \{ y_1, y_2, \dots, y_n \}$, we can derive that as long as $N_{explore}\ge\frac{1}{\log\left( 1+\frac{1}{2^n-1} \right)} \log\left( \frac{1}{\epsilon} \right)$, the assignment $AST$ can be found starting from the root node with probability at least $1-\epsilon$, i.e.,
\begin{equation}\label{eq: prob of finding AST}
    \begin{aligned}
        &\quad\ \mathbb{P}\left( \text{find assignment $AST$ } | \text{ start from root node $s_0$}\right)\\
        &= 1- \mathbb{P}\left( \text{not find assignment $AST$ } | \ s_0\right)\\
        &= 1- \mathbb{P}\left(\bigcap_{j=1}^{N_{explore}} \left\{ \text{not find $AST$ at the $j$-th exploration } \right\} |\ s_0\right)\\
        &= 1- \prod_{j=1}^{N_{explore}}\mathbb{P}\left(  \text{not find $AST$ at the $j$-th exploration } |\ s_0\right)\\
        &= 1- \prod_{j=1}^{N_{explore}} \left( 1 -  \mathbb{P}\left(  \text{find $AST$ at the $j$-th exploration } |\ s_0\right)\right)\\
        &= 1- \prod_{j=1}^{N_{explore}} \left( 1 -  \frac{n!}{\prod_{k=0}^{n-1} 2(n-k)}\right)\\
        &= 1- \left( 1 -  \frac{1}{2^n}\right)^{N_{explore}}\\
        &\ge 1-\epsilon,
    \end{aligned}
\end{equation}
where the third equation is from the independence of each exploration and the fifth equation comes with the special structures of our tree model and our uniform exploration strategy.

We have shown that our algorithm can explore each assignment $AST$ when $N_{explore}$ is sufficiently large. As a result, we have
\begin{equation}
    \label{eq: maximum converge}
    \lim_{N_{explore} \to \infty} {R}^{a_i}_{(N_i)} = \lim_{N_{i} \to \infty} {R}^{a_i}_{(N_i)} =\lim_{N_{i} \to \infty} \max\left\{{R}^{a_i}_1, {R}^{a_i}_2,\dots, {R}^{a_i}_{N_i}\right\} = {R}^{a_i}_*,\quad a.s.
\end{equation}
where ${R}^{a_i}_*$ is the maximum reward which can be attained by taking action $a_i$. Denote $\mathrm{ES} := \{ \bar{R}^{a_1},\bar{R}^{a_2},\dots, \bar{R}^{a_{|\mathcal{A}|}} \}$ and the maximum statistics sequence $\mathrm{MS} := \{ R_{(N_1)}^{a_1},R_{(N_2)}^{a_2},\dots, R_{(N_{|\mathcal{A}|})}^{a_{|\mathcal{A}|}} \}$. Then, we have
\begin{equation}\label{eq: limit of sig}
    \lim_{N_{explore}\to\infty} \operatorname{Sig}(\bar{R}^{a_i})=
    \begin{cases}
    \begin{array}{l}
        \mathbb{E}\left[ {R}^{a_i}_j \right],\ \operatorname{Rank}_{\mathrm{ES}}(\bar{R}^{a_i}) = \operatorname{Rank}_{\mathrm{MS}}({R}^{a_i}_{(N_i)})\\
        {R}^{a_i}_*,\ \operatorname{Rank}_{\mathrm{ES}}(\bar{R}^{a_i}) \neq \operatorname{Rank}_{\mathrm{MS}}({R}^{a_i}_{(N_i)}),\quad a.s.
    \end{array}
    \end{cases}
\end{equation}

We define the mapping $\operatorname{Proj}: ES\cup MS \longrightarrow MS$, where $\operatorname{Proj}(\bar{R}^{a_i}) = {R}^{a_i}_{(N_i)},\ \forall \bar{R}^{a_i}\in ES$ and $\operatorname{Proj}({R}^{a_i}_{(N_i)}) = {R}^{a_i}_{(N_i)},\ \forall {R}^{a_i}_{(N_i)} \in MS$. Combined with Formula (\ref{eq: limit of sig}), we have
\begin{equation}
    \label{eq: limit of proj}
    \lim_{N_{explore}\to\infty} \operatorname{Proj}\circ\operatorname{Sig}\left(\bar{R}^{a_i} \right) = {R}^{a_i}_*,\quad a.s. \quad \forall i\in \{1,2,\dots,|\mathcal{A}_s|\}.
\end{equation}
We take action $\hat{a} \in \arg \max_{a_i\in\mathcal{A}_s} \operatorname{Proj}\circ\operatorname{Sig}\left(\bar{R}^{a_i} \right)$. According to the definition of $\operatorname{Proj}$ and $\operatorname{Sig}$, we can access the child node that attains the optimal reward to execute the next iteration. We have proven that the DCSAT can find the optimal action for four types of SAT problems.

According to (\ref{eq: prob of finding AST}) and (\ref{eq: limit of proj}), as long as 
\begin{equation}
N_{explore}\ge\frac{1}{\log\left( 1+\frac{1}{2^n-1} \right)} \log\left( \frac{1}{\epsilon} \right),    
\end{equation}
the optimal assignment can be found starting from the root node with probability at least $1-\epsilon$, which indicates that the DCSAT will converge to the optimal solution with the highest reward when $N_{explore}$ is sufficiently large. 
\end{proof}

The above theorem concludes that the DCSAT converges to optimality under the condition of $\alpha = 0$. 
%\textcolor{red}{Additionally, the proof of convergence to the optimal strategy is given in the literature~\cite{UCT} for the case of $\alpha = 1$.} It is also feasible to substitute them into the framework of our proof based on the upper and lower bounds of explorations of each action given in literature~\cite{UCT}. 
For other $\alpha$ values, we have conducted an ablation study in the experimental section for verification. 

\subsection{Completeness of Multiple Solutions}
The DCSAT is a random algorithm that may find different optimal solutions when executed multiple times. Theorem \ref{thm2} indicates that the DCSAT has the potential to find all optimal Boolean assignments.

\begin{thm}\label{thm2}
If we set $\alpha$ in Formula (\ref{eq:equation_exploration_Esoftmax}) as zero, DCSAT with significance operator $\operatorname{Sig}$ can find all Boolean assignments provided that algorithm executions $N_{exe}$ is sufficiently large.
\end{thm}
\begin{proof}
We use the same notations as ones in the proof of Theorem \ref{thm1}.
Set all optimal Boolean assignment as $S_{{AST}^*}=\lbrace AST^*_1,AST^*_2,...,AST^*_{|S_{{AST}^*}|}\rbrace$. When $|S_{{AST}^*}|=1$, this theorem holds by Theorem \ref{thm1}. We consider $|S_{{AST}^*}|\ge 2$ in the following proof. Similar to Formula (\ref{eq:equation_thm1_2}), we have 
\begin{equation}
\label{eq:equation_thm2_1}
\lim_{N_{explore} \to \infty} N_{AST^*}=\lim_{N_{explore} \to \infty} \frac{1}{2^n} N_{explore}=\infty, \ \forall AST^* \in S_{{AST}^*},
\end{equation}
where $N_{AST^*}$ represents the number of explorations of the optimal assignment $AST^*$.

According to Formula (\ref{eq: limit of proj}), when starting from the root node, we have 
\begin{equation}
    \label{eq: limit of opt}
    \lim_{N_{explore}\to\infty} \operatorname{Proj}\circ\operatorname{Sig}\left(\bar{R}^{AST^*} \right) = {R}^{AST^*}_* \quad a.s. \quad \forall AST^* \in S_{{AST}^*}.
\end{equation}
Therefore, we select action to ${{AST}^*}$ from $S_{{AST}^*}$ by uniform policy, i.e. 
$\mathbb{P}({{AST}^*_i})=\frac{1}{|S_{{AST}^*}|}.$ Then, we have that when 
\begin{equation}
N_{exe} \ge \frac{1}{\log \left( \frac{|S_{{AST}^*}|}{|S_{{AST}^*}|-1} \right)} \log \left( \frac{|S_{{AST}^*}|}{\epsilon} \right),    
\end{equation}
the MCTS rule can find all ${{AST}^*}\in S_{{AST}^*}$, i.e.
\begin{equation}
    \label{eq:execute number inf}
    \begin{aligned}
        & \mathbb{P}\left( \text{find all } {{AST}^*}\in S_{{AST}^*} \right)\\
        =& 1-\mathbb{P}\left( \bigcup_{i=1}^{|S_{{AST}^*}|} \left\{ \text{not find } {AST}^*_i \right\} \right)\\
        \ge& 1-\sum_{i=1}^{|S_{{AST}^*}|} \mathbb{P}\left(\text{not find } {AST}^*_i \right)\\
        =&1- |S_{{AST}^*}| \left( \frac{|S_{{AST}^*}| -1}{|S_{{AST}^*}| } \right)^{N_{exe}}\\
        \ge&1-\epsilon.
    \end{aligned}
\end{equation}
It indicates that when the number of algorithm executions $N_{exe}$ approaches infinity, each $AST^*\in S_{{AST}^*}$ can be found. Then repeating the above process, DCSAT can find all the optimal Boolean assignments.
\end{proof}

\section{Experiment}
\subsection{Datasets and Experiment Setting}
We perform experiments on SATLIB and 2016 Eleventh Max-SAT Evaluation. We extract some data from SATLIB and 2016 Eleventh Max-SAT Evaluation to test the generalization of our method. Considering that the Boolean assignments that maximizes rewards is not necessarily unique, we randomly generate some data to verify it. We sequentially choose whether the variable is negated or not with a probability of 0.5, generating random numbers uniformly between 0 and 1000 as weights for the clauses. Then we use our method to generate the optimal Boolean assignments and verify it by sat4j. It is finally verified that the Boolean assignments generated by our method are not unique. For the hyperparameter setting, we set the explorations to 7 times of clauses. Furthermore, for the choice of hyperparameter $\alpha$, we select $11$ different values using the grid search method, ultimately choosing $\alpha=0.9$ as the choice for the final experiment.

\begin{table}[htpb]
\centering
\caption{Comparison of our method with solvers in the generality of solving instances. The value represents the number of instances solved by the current solver.}
\tiny
\setlength{\belowcaptionskip}{15pt}
\resizebox{.80\columnwidth}{!}{
\begin{tabular}{c|c|cc|c}
    \hline
    \textbf{Data} & \textbf{Instances} & \textbf{sat4j} & \textbf{CCEHC2akms} & \textbf{ours}\\
    \hline
    \hline
    SATLIB & 45705 & \textcolor{blue}{45705} & 1069 & \textcolor{red}{\textbf{45705}}\\   
    \hline
    Eleventh Max-SAT Evaluation & 1166 & 10 & \textcolor{blue}{865} & \textcolor{red}{\textbf{1166}}\\
    \hline
\end{tabular}}
\label{tab:sequences}
\end{table}

\subsection{Compare with Other Existing Solvers}
We conduct experiments on SATLIB with 2016 Eleventh Max-SAT Evaluation, as shown in the Table~\ref{tab:sequences}. Our method is compared with the solver sat4j and the advanced solver CCEHC2akms in the 2016 Eleventh Max-SAT Evaluation competition. We find that sat4j is only suitable for solving the data in SATLIB. On the contrary, CCEHC2akms is only applicable to the solution of the 2016 Eleventh Max-SAT Evaluation. In contrast, our method is more general and shows excellent results on both datasets.

\begin{figure*}[t]
\centering
\includegraphics[width=.90\columnwidth]{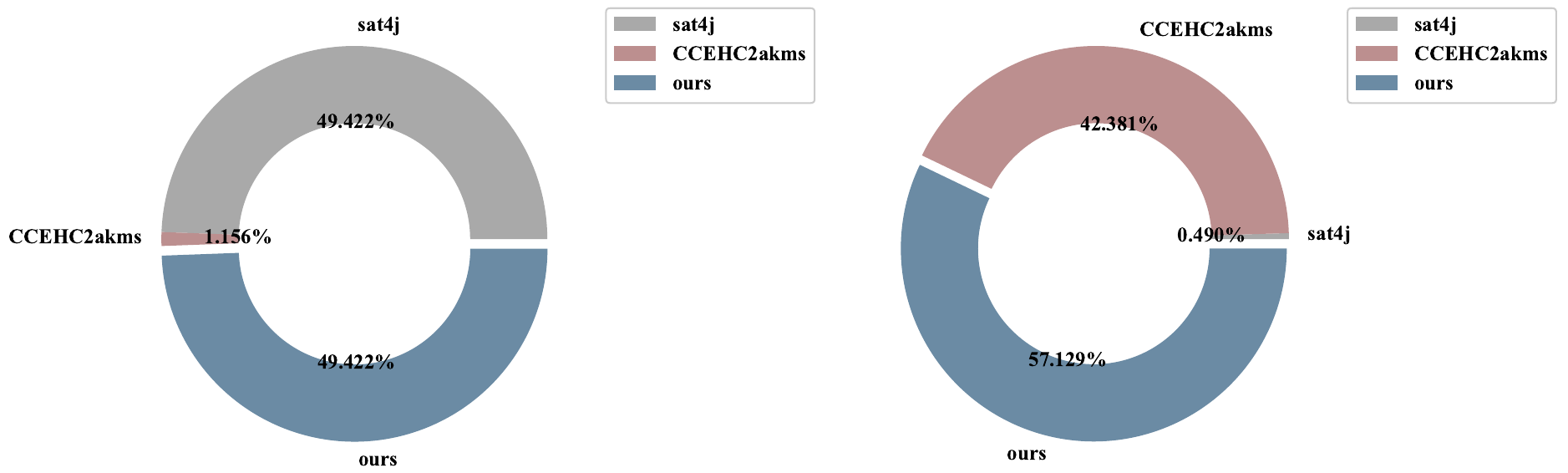} 
\caption{Visualization of the comparison of our method with the solver in terms of generality of solving instances.}
\label{fig:pie_diagram}
\end{figure*}

\begin{table}[htpb]
\centering
\caption{Presentation of results comparing our method with the solver on specific sets of instances. The value represents the number of instances solved by the current solver.}
\tiny
\setlength{\belowcaptionskip}{15pt}
\resizebox{.80\columnwidth}{!}{
\begin{tabular}{c|c|cc|c}
    \hline
    \textbf{Data} & \textbf{Instances} & \textbf{sat4j} & \textbf{CCEHC2akms} & \textbf{ours}\\
    \hline
    \hline
    uf20-91 & 1000 & \textcolor{blue}{1000} & 0 & \textcolor{red}{\textbf{1000}}\\  
    \hline
    uf50-218 & 1000 & \textcolor{blue}{1000} & 10 & \textcolor{red}{\textbf{1000}}\\
    \hline
    uf75-325 & 100 & \textcolor{blue}{100} & 0 & \textcolor{red}{\textbf{100}}\\
    \hline
    uf100-430 & 1000 & \textcolor{blue}{1000} & 4 & \textcolor{red}{\textbf{1000}}\\
    \hline
    uf125-538 & 100 & \textcolor{blue}{100} & 0 & \textcolor{red}{\textbf{100}}\\
    \hline
    uf150-645 & 100 & \textcolor{blue}{100} & 0 & \textcolor{red}{\textbf{100}}\\
    \hline
    uf175-753 & 100 & \textcolor{blue}{100} & 0 & \textcolor{red}{\textbf{100}}\\
    \hline
    RTI\_k3\_n100\_m429 & 500 & \textcolor{blue}{500} & 0 & \textcolor{red}{\textbf{500}}\\
    \hline
    BMS\_k3\_n100\_m429 & 500 & \textcolor{blue}{500} & 120 & \textcolor{red}{\textbf{500}}\\
    \hline
    Controlled Backbone Size & 40000 & \textcolor{blue}{40000} & 930 & \textcolor{red}{\textbf{40000}}\\
    \hline
    flat30-60 & 100 & \textcolor{blue}{100} & 0 & \textcolor{red}{\textbf{100}}\\
    \hline
    flat50-115 & 1000 & \textcolor{blue}{1000} & 0 & \textcolor{red}{\textbf{1000}}\\
    \hline
    flat75-180 & 100 & \textcolor{blue}{100} & 0 & \textcolor{red}{\textbf{100}}\\
    \hline
    sw100-8-lp0-c5 & 100 & \textcolor{blue}{100} & 0 & \textcolor{red}{\textbf{100}}\\
    \hline
    PHOLE & 5 & \textcolor{blue}{5} & \textcolor{blue}{5} & \textcolor{red}{\textbf{5}}\\
    \hline
    Unweighted Max-SAT & 454 & 0 & \textcolor{blue}{241}  & \textcolor{red}{\textbf{454}}\\
    \hline
    Partial Max-SAT & 210 & 10 & \textcolor{blue}{183} & \textcolor{red}{\textbf{210}}\\
    \hline
    Weighted Partial Max-SAT & 502 & 0 & \textcolor{blue}{441} & \textcolor{red}{\textbf{502}}\\
    \hline
\end{tabular}}
\label{tab:sequences}
\end{table}

For the convenience of visualization, instances solved by each solver is compared visually using a pie chart, as shown in the Figure~\ref{fig:pie_diagram}. The total instances solved by the two solvers accounted for 100\%. The ratio of instances solved by each solver to the total instances solved by two solvers is the coverage ratio of the solver in the pie chart. In this way, the proportion of solvers in Figure~\ref{fig:pie_diagram} is the proportion of instances solved by each solver. The left side of Figure~\ref{fig:pie_diagram} shows the effect on the SATLIB, and the right side shows the results on the 2016 Eleventh Max-SAT Evaluation.

More specifically, we present the specific instance categories solved on SATLIB and 2016 Eleventh Max-SAT Evaluation in the Table~\ref{tab:sequences}. Data represents the category of solved instances. Instances indicates instances for that category. We indicate the best performing solver effect in red and the second-best solver effect in blue.

\begin{figure*}[htpb]
\centering
\includegraphics[width=.7\columnwidth]{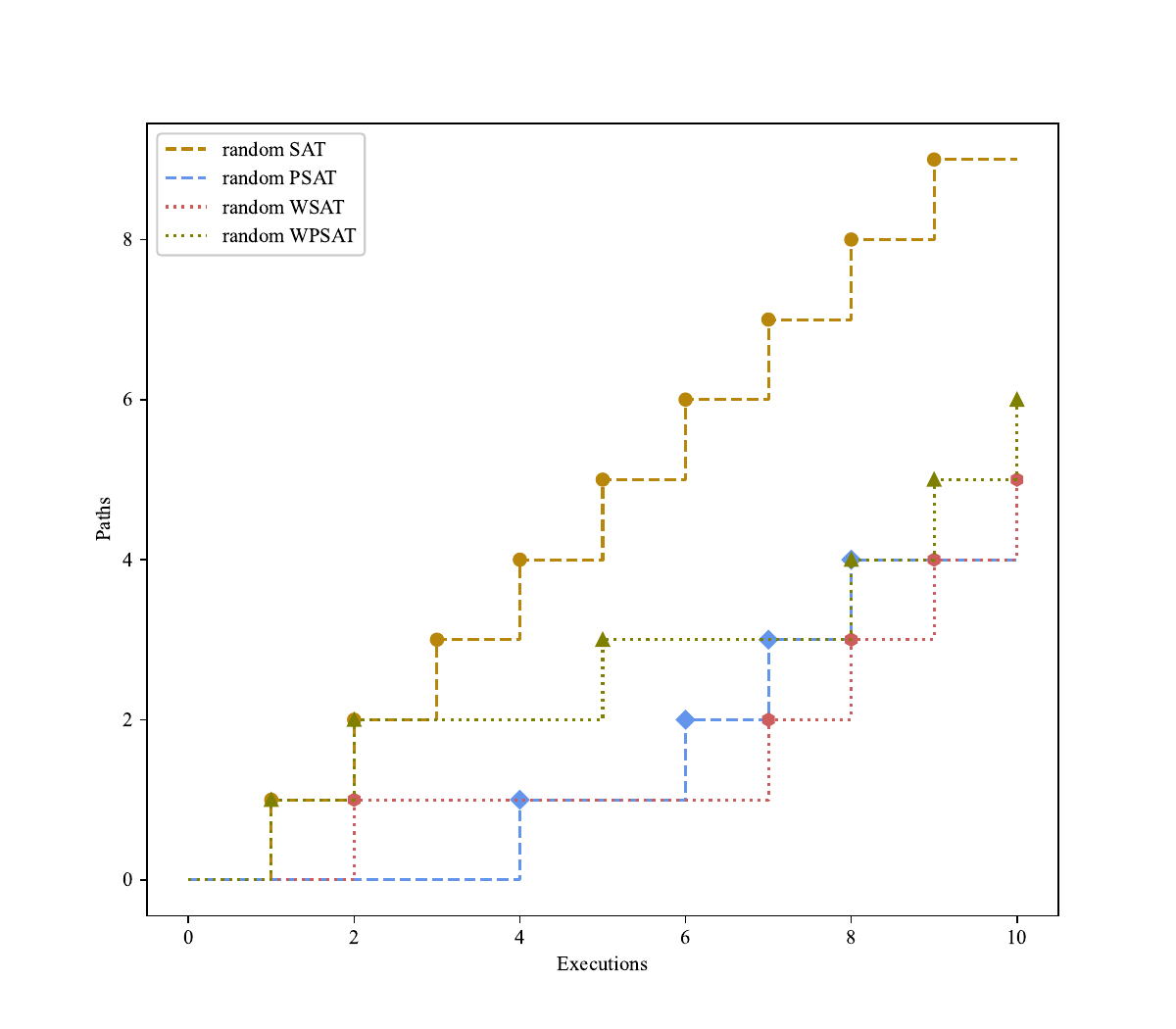} 
\caption{Multiple optimal solutions found vary with executions of our algorithm. The X-axis represents algorithm executions, and the Y-axis represents different optimal solutions currently found. }
\label{fig:multiple_paths}
\end{figure*}

\begin{table}[htpb]
\centering
\caption{Multiple Boolean assignments with maximum objective values on four randomly generated typical instances.}
\tiny
\setlength{\belowcaptionskip}{15pt}
\resizebox{1.00\columnwidth}{!}{
\begin{tabular}{c|cccc}
    \hline
    \textbf{Problem} & \textbf{Index} & \textbf{Pivot paths} & \textbf{Objective value} \\
    \hline
    \hline
    \multirow{9}{*}{random SAT} & 1 & $\left[0., 0., 1., 1., 1., 0., 0., 1., 0., 0., 0., 1., 1., 0., 0., 0., 1. \right]$ & 33 \\
    \cline{2-4}
    & 2 & $\left[0., 0., 1., 1., 1., 1., 1., 0., 0., 1., 0., 1., 0., 0., 0., 0., 1. \right]$ & 33 \\
    \cline{2-4}
    & 3 & $\left[0., 1., 1., 0., 0., 0., 0., 1., 0., 0., 0., 0., 1., 0., 0., 0., 1. \right]$ & 33 \\
    \cline{2-4}
    & 4 & $\left[0., 1., 1., 0., 0., 0., 1., 0., 0., 0., 1., 0., 0., 0., 0., 0., 0. \right]$ & 33 \\
    \cline{2-4}
    & 5 & $\left[1., 0., 0., 1., 0., 0., 0., 0., 0., 0., 0., 1., 0., 0., 0., 0., 0. \right]$ & 33 \\
    \cline{2-4}
    & 6 & $\left[1., 0., 0., 1., 0., 0., 0., 1., 1., 1., 0., 0., 0., 0., 0., 0., 0. \right]$ & 33 \\
    \cline{2-4}
    & 7 & $\left[1., 0., 0., 1., 0., 0., 1., 0., 0., 1., 0., 0., 0., 1., 0., 0., 0. \right]$ & 33 \\
    \cline{2-4}
    & 8 & $\left[1., 0., 0., 1., 0., 1., 1., 0., 1., 0., 0., 1., 0., 0., 0., 0., 0. \right]$ & 33 \\
    \cline{2-4}
    & 9 & $\left[1., 0., 0., 1., 1., 0., 1., 0., 1., 1., 0., 0., 0., 1., 0., 0., 0. \right]$ & 33 \\
    \cline{2-4}
    \hline
    \hline

    \multirow{4}{*}{random PSAT} & 1 & $\left[1., 1., 1., 1., 0., 1., 0., 1. \right]$ & 6031 \\
    \cline{2-4}
    & 2 & $\left[1., 1., 1., 1., 1., 1., 0., 1. \right]$ & 6031 \\
    \cline{2-4}
    & 3 & $\left[0., 1., 1., 1., 0., 0., 0., 0. \right]$ & 6031 \\
    \cline{2-4}
    & 4 & $\left[0., 1., 1., 1., 0., 1., 0., 0. \right]$ & 6031 \\
    \cline{2-4}
    \hline
    \hline

    \multirow{5}{*}{random WSAT} & 1 & $\left[1., 1., 1., 0., 0., 0., 1., 1., 1., 1., 0. \right]$ & 28627 \\
    \cline{2-4}
    & 2 & $\left[1., 1., 1., 0., 1., 0., 0., 1., 0., 1., 1. \right]$ & 28627 \\
    \cline{2-4}
    & 3 & $\left[1., 1., 1., 0., 1., 0., 0., 1., 0., 1., 0. \right]$ & 28627 \\
    \cline{2-4}
    & 4 & $\left[1., 1., 1., 0., 1., 0., 1., 1., 0., 1., 0. \right]$ & 28627 \\
    \cline{2-4}
    & 5 & $\left[1., 1., 1., 1., 0., 0., 1., 1., 1., 1., 0. \right]$ & 28627 \\
    \cline{2-4}
    \hline
    \hline

    \multirow{6}{*}{random WPSAT} & 1 & $\left[0., 0., 0., 1., 0., 0., 0., 0., 1., 0., 0., 0., 0., 0., 1., 0., 0., 0., 0., 0. \right]$ & 718406 \\
    \cline{2-4}
    & 2 & $\left[0., 0., 0., 1., 1., 0., 0., 0., 1., 0., 0., 0., 0., 0., 1., 1., 0., 1., 0., 1. \right]$ & 718406 \\
    \cline{2-4}
    & 3 & $\left[0., 1., 0., 1., 1., 1., 1., 0., 1., 1., 0., 0., 0., 0., 1., 1., 0., 0., 1., 0. \right]$ & 718406 \\
    \cline{2-4}
    & 4 & $\left[0., 1., 0., 1., 1., 1., 1., 1., 1., 1., 0., 0., 0., 1., 0., 1., 0., 0., 0., 0. \right]$ & 718406 \\
    \cline{2-4}
    & 5 & $\left[0., 1., 1., 1., 0., 0., 0., 1., 1., 0., 0., 0., 0., 0., 0., 1., 0., 1., 1., 0. \right]$ & 718406 \\
    \cline{2-4}
    & 6 & $\left[1., 1., 1., 1., 0., 0., 1., 0., 1., 0., 1., 0., 0., 0., 1., 0., 0., 0., 0., 1. \right]$ & 718406 \\
    \cline{2-4}
    \hline
    \hline
\end{tabular}}
\label{tab:sequences}
\end{table}

\subsection{Findings of Multiple Boolean Assignments}
In our reinforcement learning model, performed actions are guided by rewards. Consider that the Boolean assignment that maximizes the reward is not unique. Therefore, the obtained results are not unique in the multiple executions of the algorithm. We prove theoretically that all optimal Boolean assignments can be found as executions of the algorithm tends to infinity. We verify that our method can find different Boolean assignments on four types of randomly generated instances. We perform each of the randomly generated instances $10$ times. It is found that different Boolean assignments can be obtained, as shown in the Table~\ref{tab:sequences}. Furthermore, we visualize the different assignments previously discovered versus algorithm executions, as shown in Figure~\ref{fig:multiple_paths}. It can be found that as algorithm executions increases, our method can find different Boolean assignments.

\begin{table}[htpb]
\centering
\caption{Additional comparison models constructed for the SAT problem. Model 1, Model 2 and Model 3 are newly constructed comparison models. Model Initial indicates the original model.}
\tiny
\resizebox{1\columnwidth}{!}{
\begin{tabular}{ccccc}
    \hline
    \textbf{Model} & \textbf{State} & \textbf{Action} & \textbf{Reward} \\
    \hline
    Model 1 & SAT tableaux & $A_{1}=A-\{a\}$ & $R_{1}=\sum_{i=1}^{n-1} w_{i}^1 (c^T y_{i+1}-c^T y_{i})$ \\
    Model 2 & SAT tableaux & $A_{1}=A-\{a\}$ & $R_{2}=\sum_{i=1}^{n} w_{i}^2 (c^T y_{i})$ \\
    Model 3 & SAT tableaux & $A_{1}=A-\{a\}$ & $R_{4}=0.5 R_{2}+0.5 R_{3}$ \\
    Model Initial & SAT tableaux & $A_{1}=A-\{a\}$ & $R_{3}=c^T y^*$ \\
    \hline
\end{tabular}}
\label{tab:model}
\end{table}

\subsection{Ablation Study}
Inspired by MCTS rule, we consider three approximations based rewards to compare whether they can improve the experimental effect. Our Initial model is denoted as Model Initial, and the models obtained by embedding the three kinds of rewards are denoted as Model 1, Model 2 and Model 3 respectively. The four comparison models are shown in Table~\ref{tab:model}.

\begin{figure}[htpb]
    \centering
    \subfigure[\label{fig:a}CBS\_k3\_n100\_m435\_b10\_14]
    {\includegraphics[width=1.00\columnwidth]{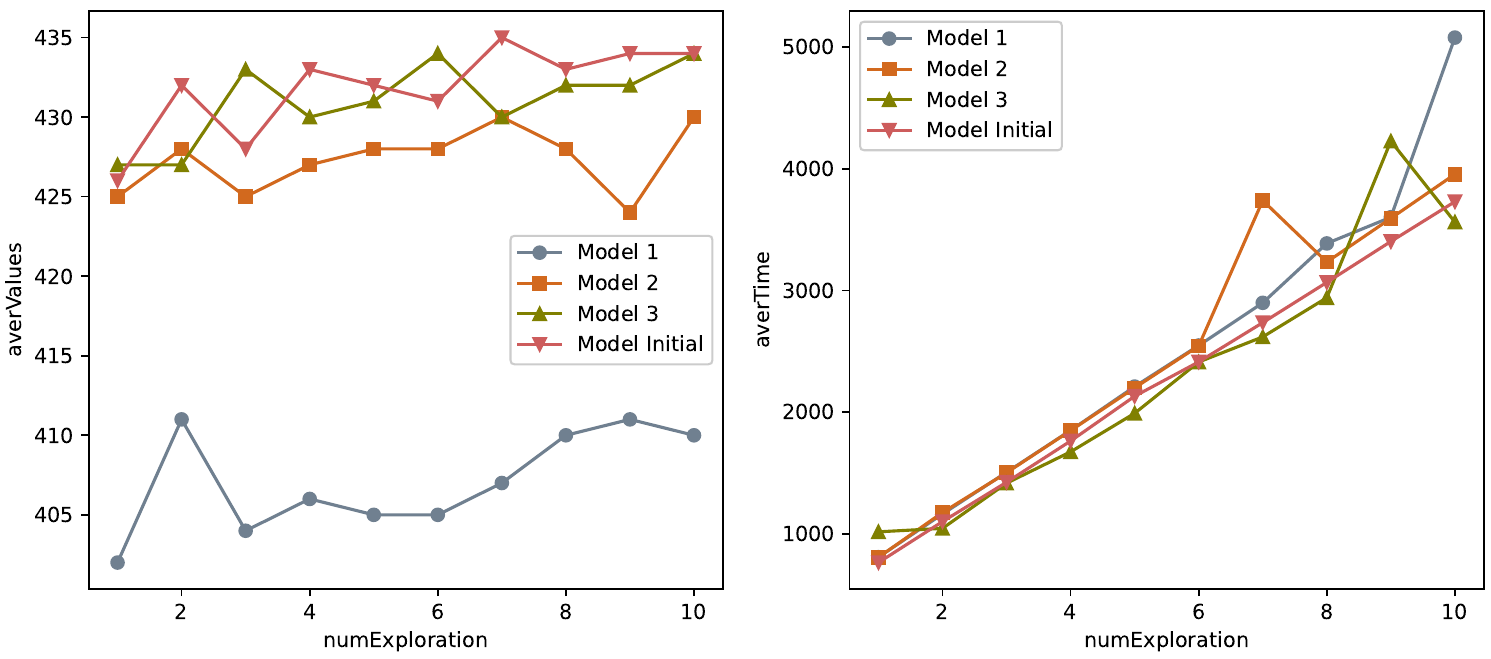}}
    \vspace{0.5em}
 
    \centering
    \subfigure[\label{fig:b}RTI\_k3\_n100\_m429\_25]
    {\includegraphics[width=1.00\columnwidth]{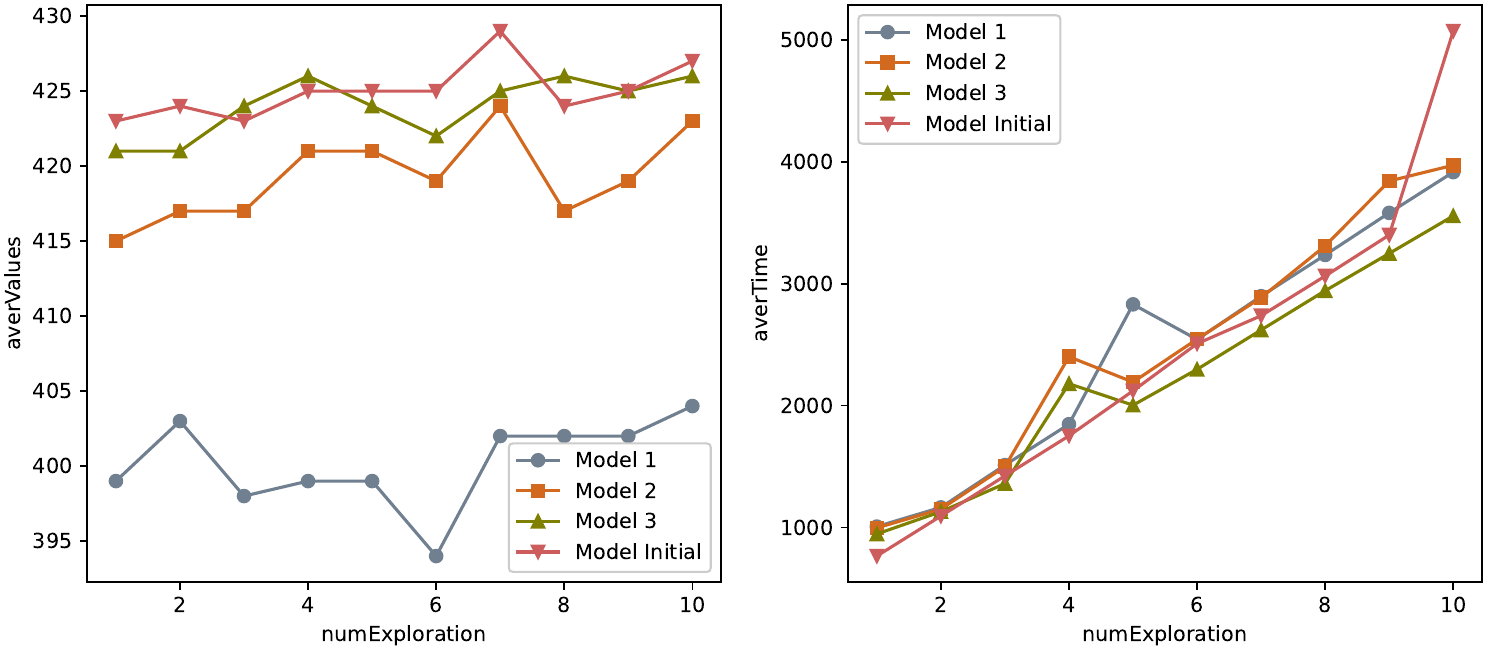}}
    \vspace{0.5em}
 
    \centering
    \subfigure[\label{fig:c}uf125-052]
    {\includegraphics[width=1.00\columnwidth]{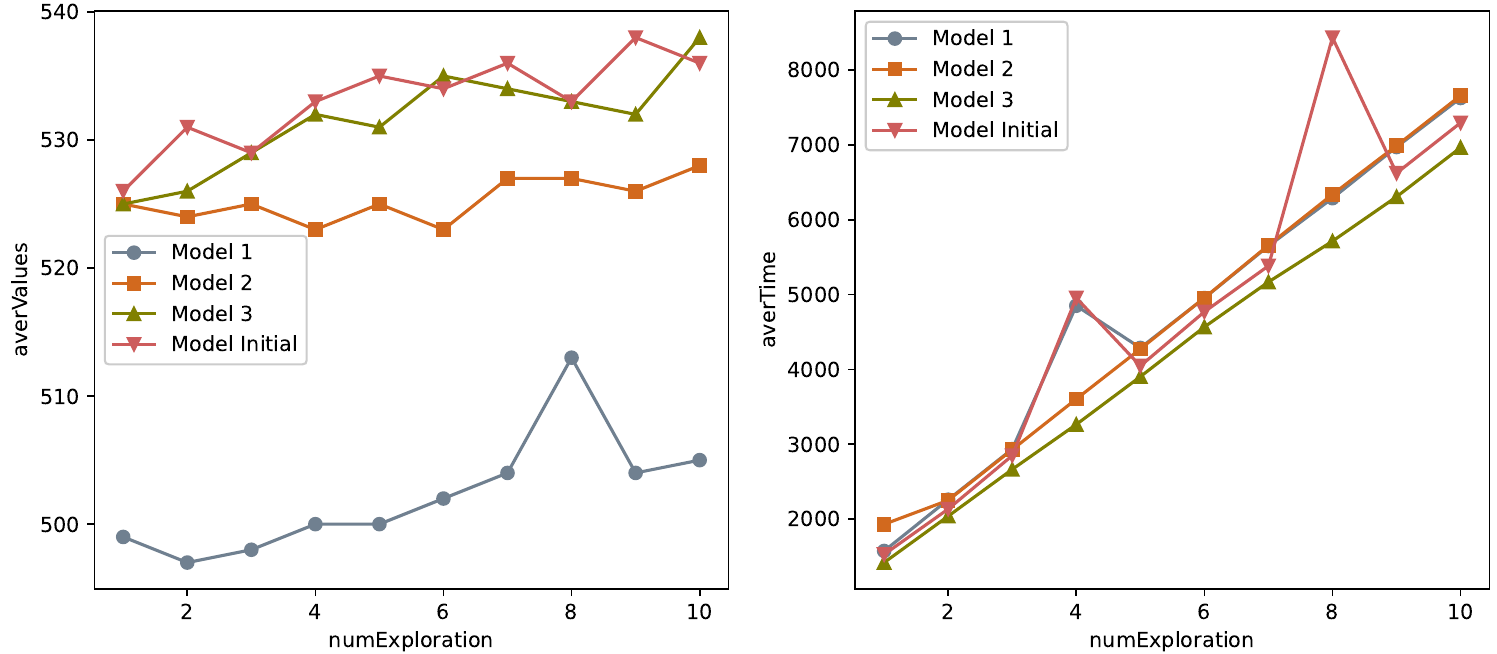}}
    \vspace{0.5em}
\end{figure}

\begin{figure}[htpb]
    \centering
    \subfigure[\label{fig:d}s3v70c900-2]
    {\includegraphics[width=1.00\columnwidth]{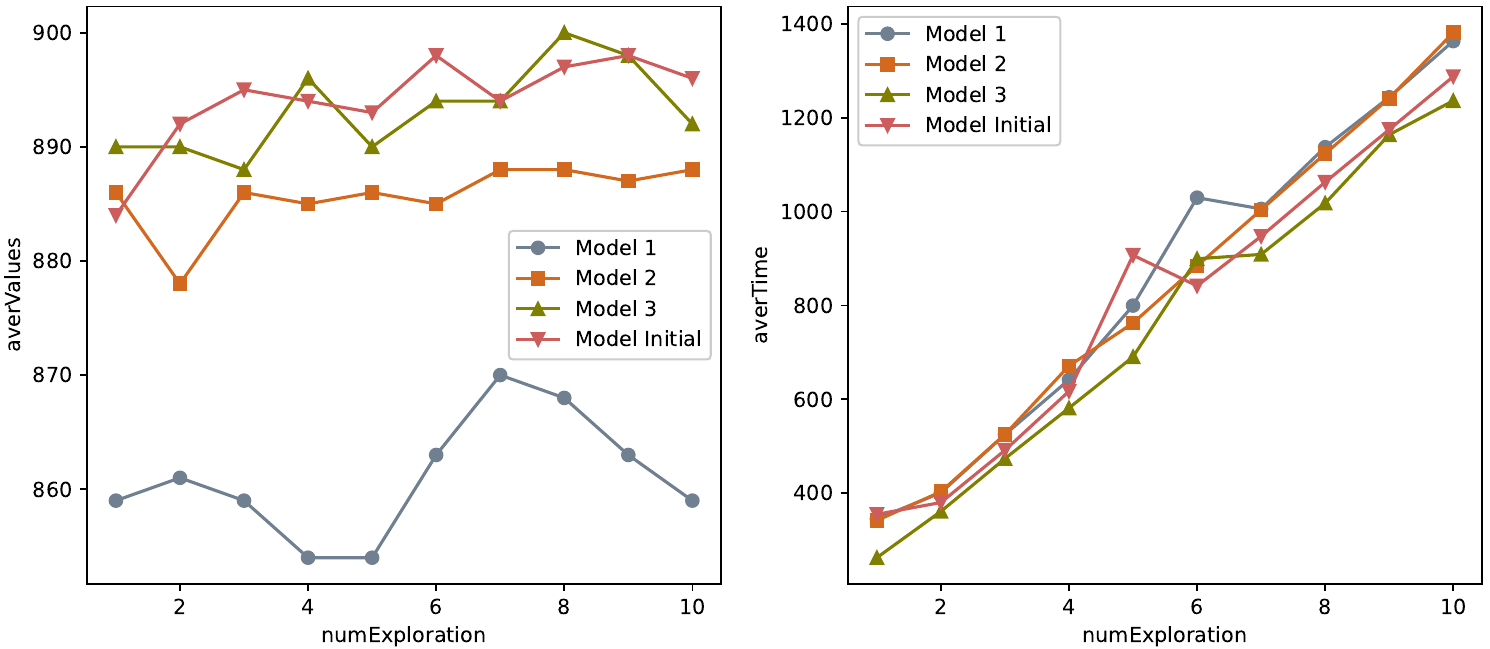}}
    \vspace{0.5em}
 
    \centering
    \subfigure[\label{fig:e}file\_rpms\_wcnf\_L3\_V100\_C800\_H100\_1]
    {\includegraphics[width=1.00\columnwidth]{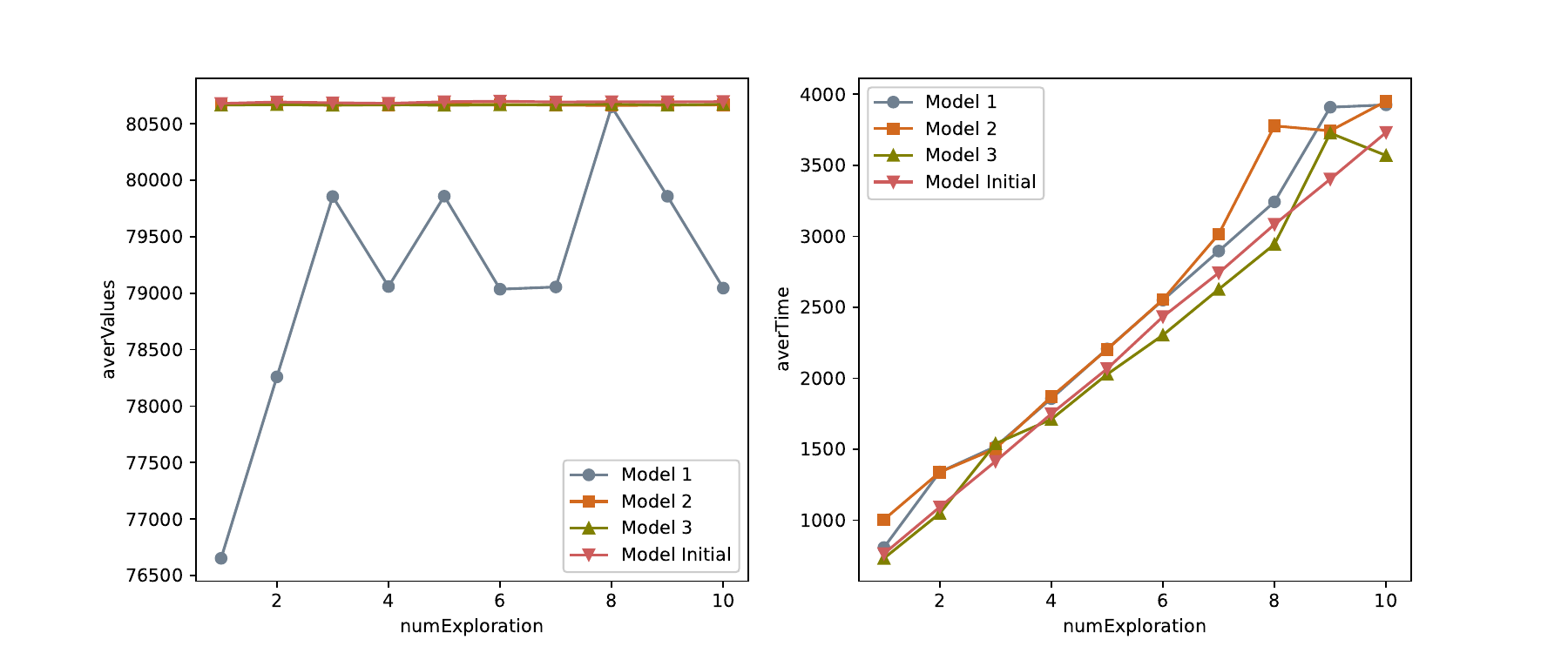}}
    \vspace{0.5em}
 
    \centering
    \subfigure[\label{fig:e}file\_rwpms\_wcnf\_L3\_V100\_C800\_H100\_6]
    {\includegraphics[width=1.00\columnwidth]{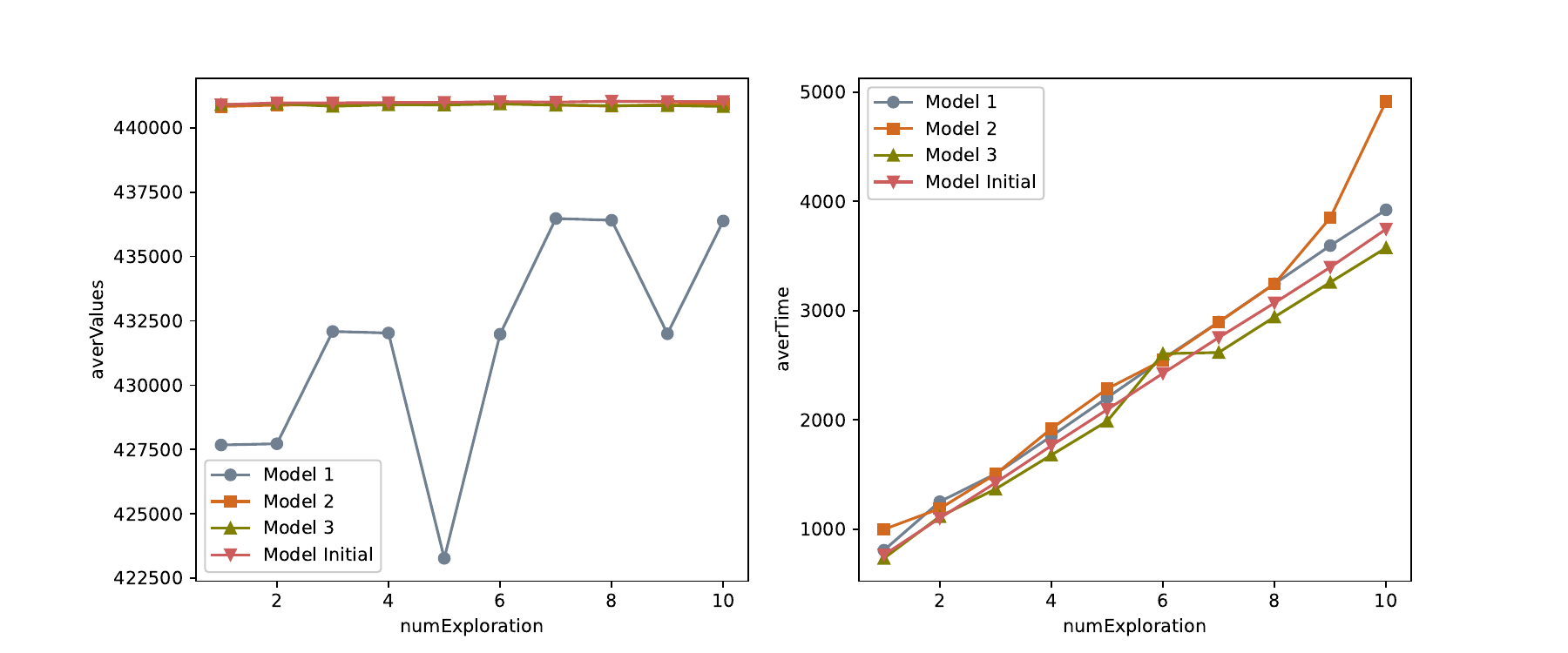}}
    \vspace{0.5em}
    \caption{Model comparison figure on six representative instances. The X-axis represents the explorations, which are multiples of Boolean variables. The Y-axis of the left figure represents the average objective values. And the Y-axis of the right figure represents the average solution time.}
    \label{fig:Model_ititalInstances}
\end{figure}

The first reward is a weighted sum of objective value increments, as shown in Formula (\ref{eq:reward_1}).
\begin{equation}
\label{eq:reward_1}
R_{1}=\sum_{i=1}^{n-1} w_{i}^1 (c^T y_{i+1}-c^T y_{i}) \ ,\ \ 
w_{i}^1=\frac{n-i}{n}.
\end{equation}
Where $n$ represents the number of binary variables, $i$ represents the $i^{th}$ assignment, $y_{i}$ is the locally feasible solution obtained from the $i^{th}$ assignment, and $w_{i}^1 \in (0,1]$ is the weight. It is noteworthy that the proposed linear weight factor provides the weight of linear attenuation according to the order of assignments. The second reward is a linearly weighted sum of the objective values, as shown in Formula (\ref{eq:reward_2}). $w_{i}^2 \in (0,1]$ is also a linearly decaying weight, similar to the definition of $w_{i}^1$. Different from $R_1$ focuses on the increment of objective values, $R_2$ is more inclined to objective values brought by the earlier assignment.
\begin{equation}
\label{eq:reward_2}
R_{2}=\sum_{i=1}^{n} w_{i}^2 (c^T y_{i}) \ ,\ \ 
w_{i}^2=\frac{(n+1)-i}{n}.
\end{equation}
Formula (\ref{eq:reward_4}) defines the third reward, which is the equally weighted sum of $R_2$ and $R_3$. While emphasizing the importance of initial assignment, diversity of exploration paths is also encouraged.
\begin{equation}
\label{eq:reward_4}
R_{3}=0.5 R_{2}+0.5 R_{3}.
\end{equation}

\begin{figure*}[htbp]
    \centering
    \subfigure[\label{fig:a}CBS\_k3\_n100\_m435\_b10\_14.cnf]  % (a)小标题 
    {  
        \begin{minipage}[t]{0.50\linewidth}  % 左图分配70%
        \centering
        \includegraphics[width=1.00\columnwidth]{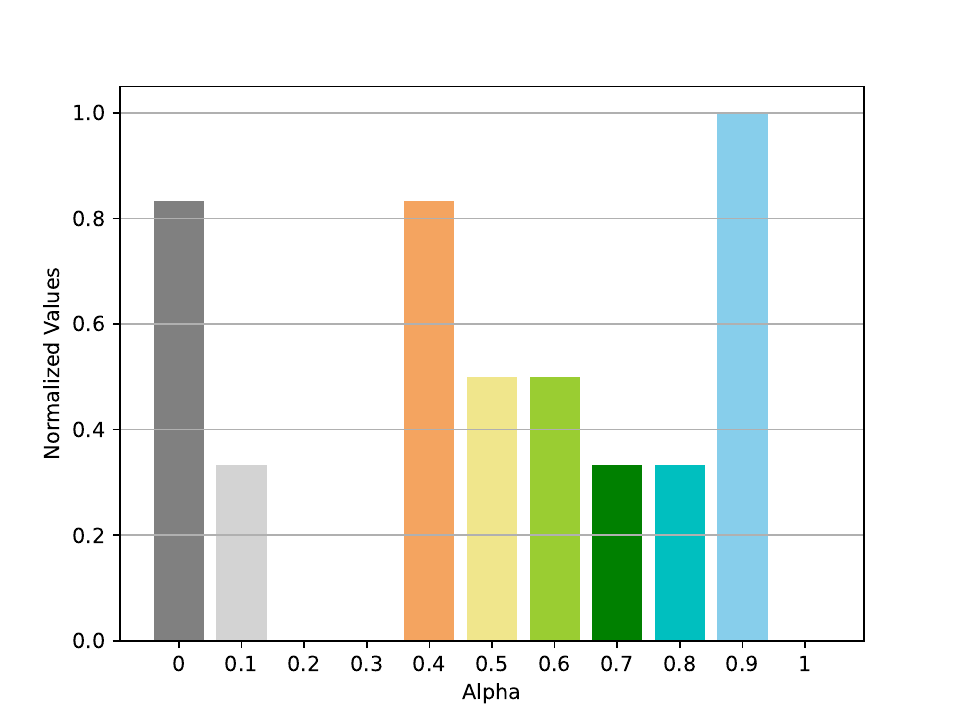}
        %\caption{fig1}
        \end{minipage}%
    }%
    \subfigure[\label{fig:b}RTI\_k3\_n100\_m429\_25.cnf]  % (a)小标题 
    {
        \begin{minipage}[t]{0.5\linewidth}  % 右图分配30%
        \centering
        \includegraphics[width=1.00\columnwidth]{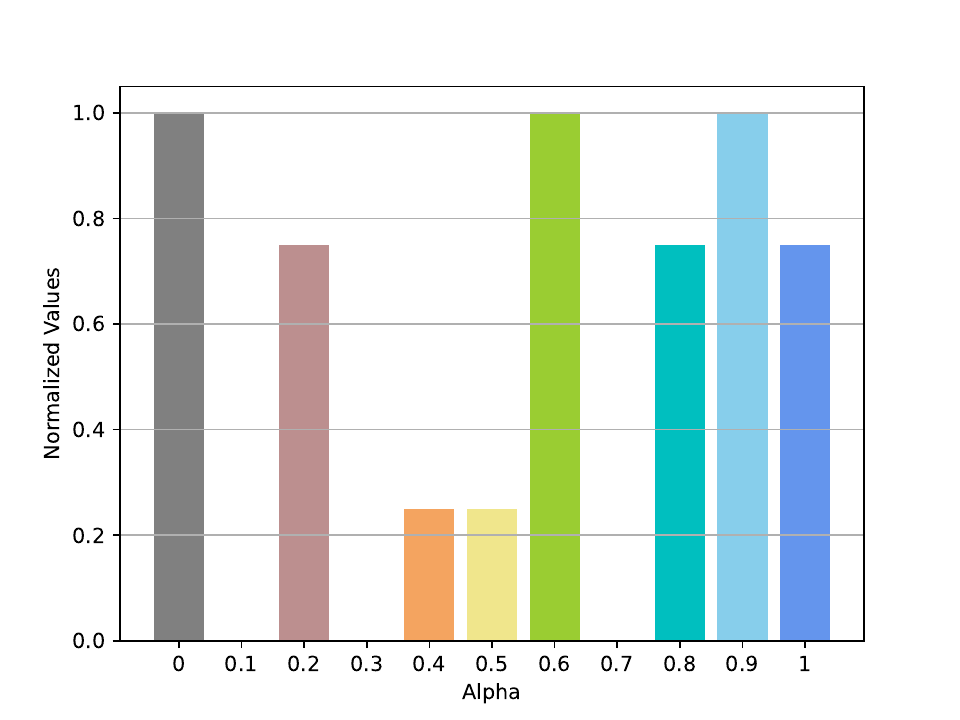}
        % \caption{fig2}
        \end{minipage}%
    }%
    
    \centering
    \subfigure[\label{fig:c}uf125-052.cnf]  % (a)小标题 
    {  
        \begin{minipage}[t]{0.50\linewidth}  % 左图分配70%
        \centering
        \includegraphics[width=1.00\columnwidth]{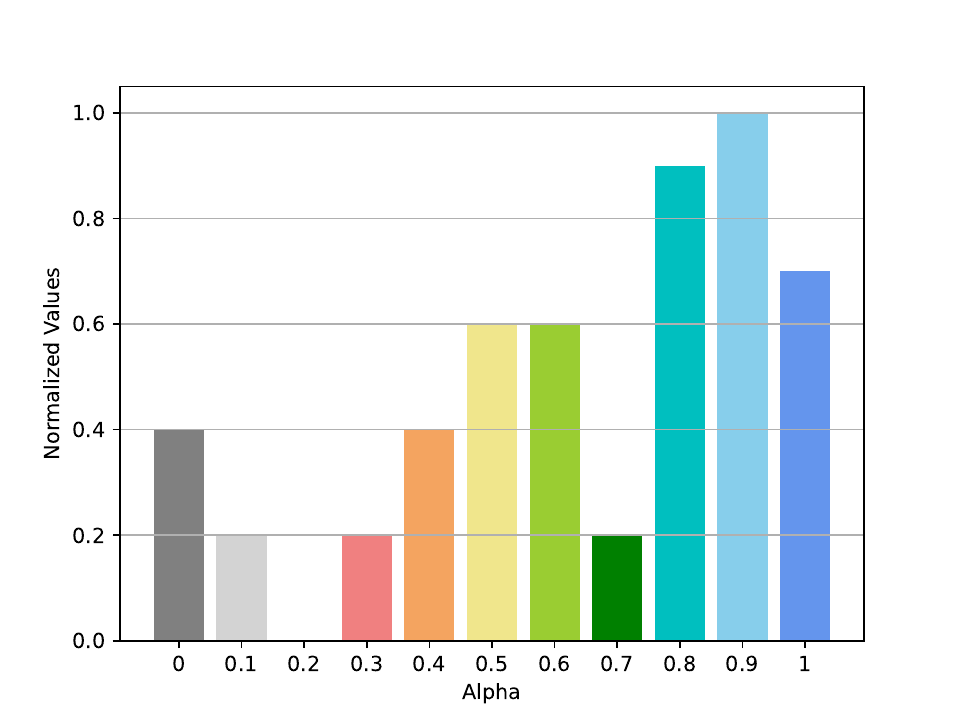}
        %\caption{fig1}
        \end{minipage}%
    }%
    \subfigure[\label{fig:c}file\_rpms\_wcnf\_L3\_V100\_C800\_H100\_1.wcnf]  % (a)小标题 
    {  
        \begin{minipage}[t]{0.50\linewidth}  % 左图分配70%
        \centering
        \includegraphics[width=1.00\columnwidth]{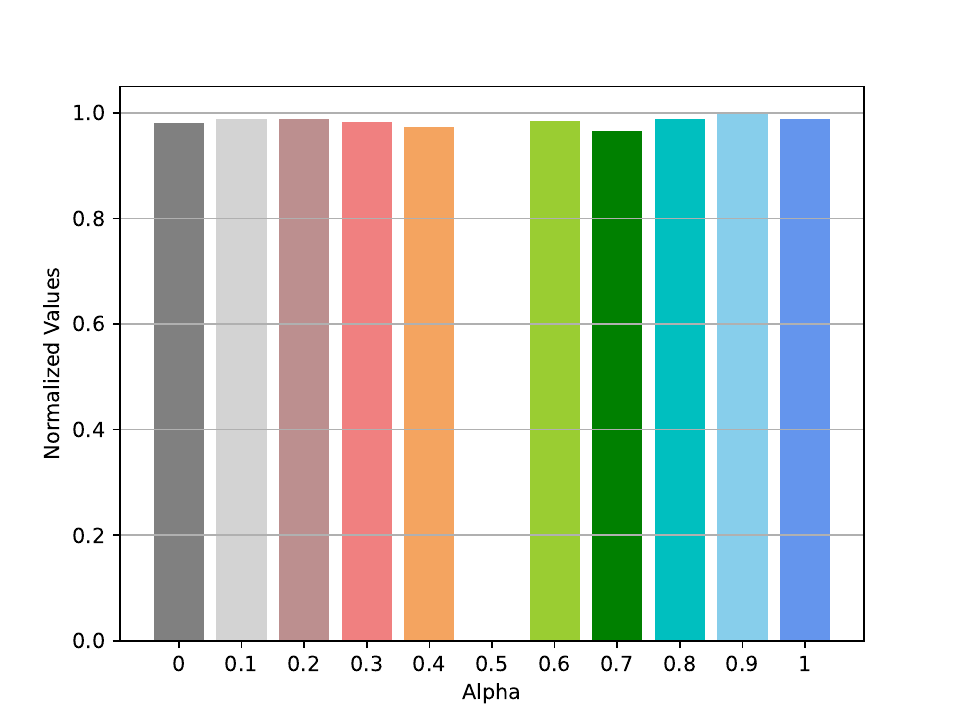}
        %\caption{fig1}
        \end{minipage}%
    }%

    \centering
    \subfigure[\label{fig:d}s3v70c900-2.cnf]  % (a)小标题 
    {
        \begin{minipage}[t]{0.5\linewidth}  % 右图分配30%
        \centering
        \includegraphics[width=1.00\columnwidth]{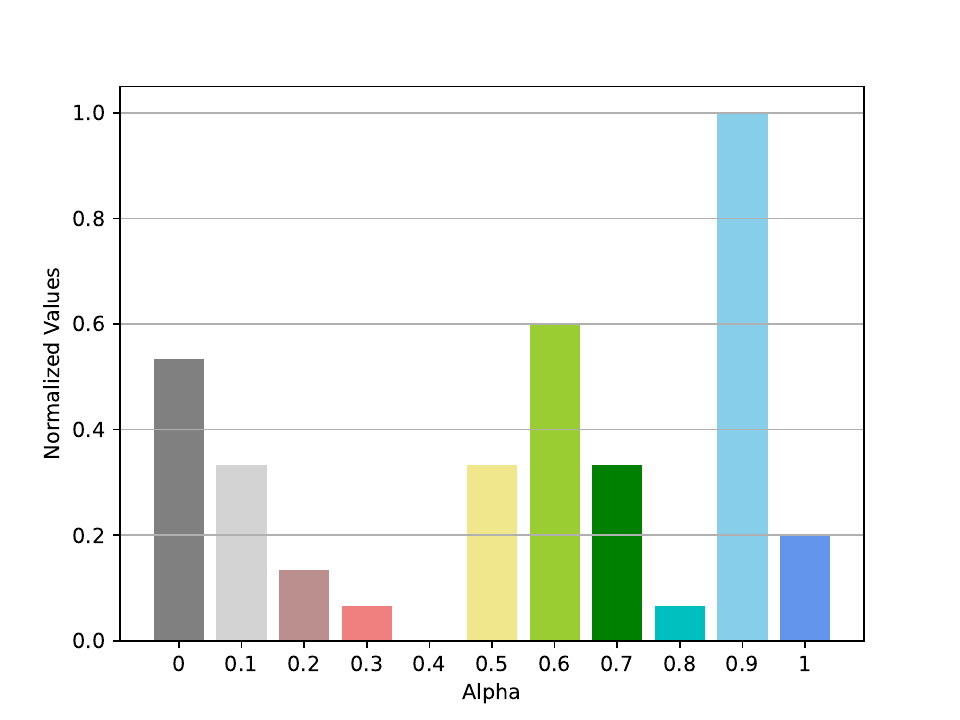}
        % \caption{fig2}
        \end{minipage}%
    }%
    \subfigure[\label{fig:d}file\_rwpms\_wcnf\_L3\_V100\_C800\_H100\_6.wcnf]  % (a)小标题 
    {
        \begin{minipage}[t]{0.5\linewidth}  % 右图分配30%
        \centering
        \includegraphics[width=1.00\columnwidth]{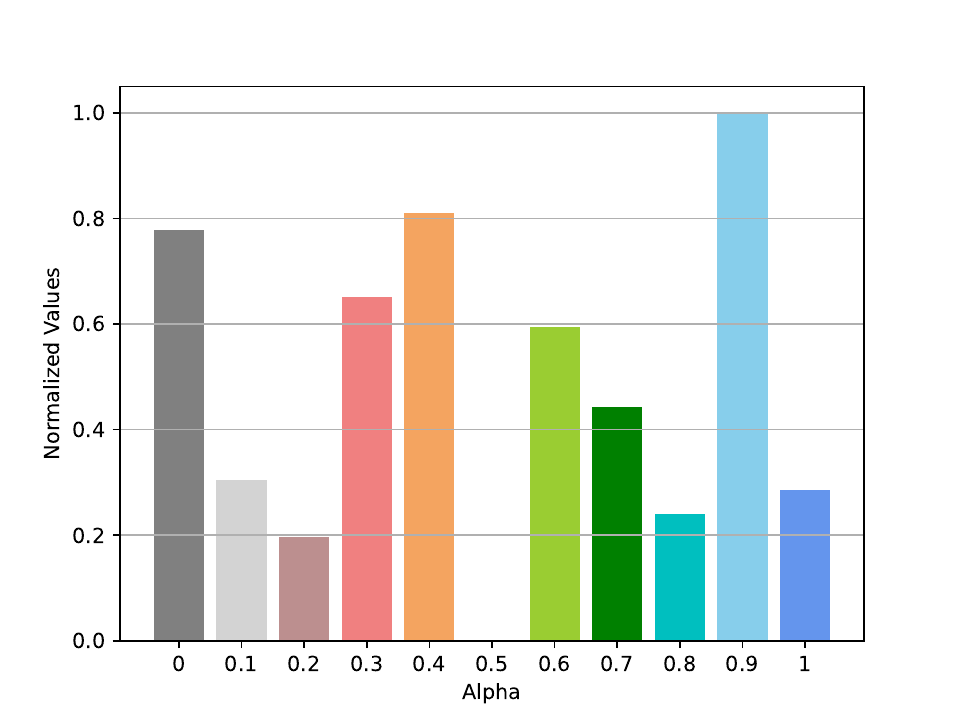}
        % \caption{fig2}
        \end{minipage}%
    }%
\caption{Relationship between the average number of pivot iterations and $\alpha$ under different initial explorations. The X-axis represents different problems, and the Y-axis represents the average pivot iterations. }  % 标题
\label{fig:Alpha}  % 标签
\end{figure*}

We conduct comparative experiments on $6$ typical instances, as shown in the Figure~\ref{fig:Model_ititalInstances}. Each subgraph of Figure~\ref{fig:Model_ititalInstances} represents the experimental effect on a typical instance. We use different color line represent different models. The left side of the subfigures shows relationship between average objective values and explorations. It can be seen from the left subfigure that the average objective value of Model Initial is the best and the optimal value can be reached. The right side shows the relationship between computation time and explorations. Similarly, the right subfigure shows that Model 3 has the shortest average computation time, and Model Initial is not much different from it. Considering the computation time and solution quality, Model Initial has the best overall performance.

We also conduct a comparative experiment to determine the best value of $alpha$, as shown in the Figure~\ref{fig:Alpha}. Using the grid search method, we divide $alpha$ from $0$ to $1$ into $11$ equal parts at $0.1$ intervals. We follow the six representative data sets selected previously. We solve each problem $20$ times and record its objective value. Given the wide range of objective values, we use the interquartile range normalization to normalize them to between zero and one. From the results of six typical instances, it can be seen that the penultimate cylinder performs the best. Therefore, we conclude that $alpha$ is appropriate for a value of $0.9$, which is in line with our intuitive understanding of soft max.

\section{Conclusion}
We propose a unified solving paradigm for four types of SAT problems (MaxSAT, Weighted MaxSAT, PMS and WPMS). Notably, our method transforms four types of SAT problems into general 0-1 integer programming structures by subtly adjusting weights. Consolidated reinforcement learning models are subsequently proposed to make this approach applicable to the reinforcement learning paradigm. In this way, the Monte carlo tree search algorithm can effectively evaluate the pros and cons of each variable Boolean value assignment, and sharply cut off unnecessarily search space and find solution of the problem. Depending on the randomness of the algorithm, different Boolean assignments can be obtained when the algorithm is executed repeatedly. Theoretical derivation and experimental verification demonstrate the superiority of the proposed framework. To some extent, it enhances the diversity of labels for supervised learning methods. This idea also applies to other combinatorial optimization problems that can be transformed into tree search structures. In the future, we also plan to approximate this framework with deep neural networks combined with reinforcement learning to achieve more efficient time efficiency.

%\section*{CRediT authorship contribution statement} 
%\textbf{Anqi Li:} Conceptualization, Methodology, Experiment, Writing - original draft. \textbf{Bonan Li:} Experiment, Validation. \textbf{Congying Han:} Supervision, Writing - review \& editing. \textbf{Tiande Guo:} Supervision, Writing - review \& editing. 

%\section*{Declaration of competing interest} The authors declare that they have no known competing financial interests or personal relationships that could have appeared to influence the work reported in this paper.

\section*{Acknowledgements} This paper is supported by the National Key R\&D Program of China [grant number 2021YFA1000403]; the National Natural Science Foundation of China [grant number 11991022];  the Strategic Priority Research Program of Chinese Academy of Sciences [grant number XDA27000000]; and the Fundamental Research Funds for the Central Universities.

%\section*{References}

\bibliography{mybibfile}

\end{document}